%% file: 3dv_review_main.tex
\documentclass[10pt,twocolumn,letterpaper]{article}

\usepackage{3dv}
\usepackage{times}
\usepackage{epsfig}
\usepackage{graphicx}
\usepackage{amsmath}
\usepackage{amssymb}

\usepackage{comment}
\usepackage{subcaption}
\usepackage{multirow}
\usepackage{upgreek}
\usepackage{xspace}
\usepackage[dvipsnames]{xcolor}
\usepackage{multirow}
\usepackage{tabularx}

\usepackage[pagebackref=true,breaklinks=true,letterpaper=true,colorlinks,bookmarks=false]{hyperref}

\newcommand{\SfM}{S\textit{f}M\xspace}
\newcommand{\SfC}{S\textit{f}C\xspace}
\def\SfMpp{S\textit{f}M++\xspace}

\threedvfinalcopy % *** Uncomment this line for the final submission

\def\threedvPaperID{205} % *** Enter the 3DV Paper ID here

\ifthreedvfinal\fi
\begin{document}

%%%%%%%%% TITLE
\title{Deep NR\SfMpp: Towards Unsupervised 2D-3D Lifting in the Wild}
% \\ \ch I don't think indictive biases is correct, as the priors baked in to the architecture is not inductive. It's not induced from previous examples
% }

\author{Chaoyang Wang$^1$~~~~Chen-Hsuan Lin$^1$~~~~Simon Lucey$^{1,2}$\\
$^1$Carnegie Mellon University~~~~$^2$Argo AI, LLC\\
{\tt\small \{chaoyanw, chlin, slucey\}@cs.cmu.edu}}
\maketitle

\input{notation.tex}

%%%%%%%%% ABSTRACT
\input{sections/0-abstract.tex}

\input{sections/1-intro.tex}
\input{sections/2-relatedwork.tex}

\input{sections/3-method.tex}

\input{sections/4-deepnrsfm++.tex}

\input{sections/5-experiment.tex}

\input{sections/6-conclusion.tex}

\input{sections/supp}

{\small
\bibliographystyle{ieee}
\bibliography{egbib}
}

\end{document}

%% file: notation.tex
\def\cI{\mathcal{I}}
\def\cT{\mathcal{T}}
\def\p{\mathbf{p}}
\def\t{\mathbf{t}}
\def\R{\mathbf{R}}
\def\D{\mathbf{D}}
\def\A{\mathbf{A}}
\def\B{\mathbf{B}}
\def\I{\mathbf{I}}
\def\supi{{(i)}}
\def\bd{\mathbf{d}}
\def\bz{\mathbf{z}}
\def\bw{\mathbf{w}}
\def\M{\mathbf{M}}
\def\S{\mathbf{S}}
\def\W{\mathbf{W}}
\def\w{\mathbf{w}}
\def\x{\mathbf{x}}
\def\t{\mathbf{t}}
\def\cW{\mathcal{W}}
\def\cL{\mathcal{L}}
\def\bvarphi{\boldsymbol{\varphi}}
\def\btheta{\boldsymbol{\theta}}
\def\blambda{\boldsymbol{\lambda}}
\def\bPsi{\mathbf{\Psi}}
\def\bpsi{\boldsymbol{\psi}}

\def\s{\mathbf{s}}
\def\Real{\mathbb{R}}
\def\so{\mathfrak{so}}
\def\SO{\mathbb{SO}}
\def\1{\mathbf{1}}
\def\xy{\text{xy}}
\def\Z{\mathbf{Z}}
\def\d{\mathbf{d}}
\def\C{\mathcal{C}}
\def\tW{\widetilde{\W}}
\def\tD{\widetilde{\D}}
\def\td{\widetilde{\d}}
\def\tbPsi{\widetilde{\bPsi}}
\def\tP{\widetilde{P}}

%% file: sections/0-abstract.tex
\begin{abstract}

The recovery of 3D shape and pose from 2D landmarks stemming from a large ensemble of images can be viewed as a non-rigid structure from motion (NR\SfM) problem.
Classical NR\SfM approaches, however, are problematic as they rely on heuristic priors on the 3D structure (\eg low rank) that do not scale well to large datasets.
% To date, however, the application of NR\SfM to problems in the wild has been problematic.
% Classical NR\SfM approaches do not scale to large {\edit image datasets} and can only handle certain types of 3D structure (\eg low rank).
Learning-based methods are showing the potential to reconstruct a much broader set of 3D structures than classical methods -- dramatically expanding the importance of NR\SfM to atemporal unsupervised 2D to 3D lifting. Hitherto, these learning approaches have not been able to effectively model perspective cameras or handle missing/occluded points -- limiting their applicability to in-the-wild datasets.
In this paper, we present a generalized strategy for improving learning-based NR\SfM methods~\cite{ck19} to tackle the above issues.
% We demonstrate how a recent deep factorization method~\cite{ck19} can take advantage of this generalization.
Our approach, \emph{Deep NR\SfMpp}, achieves state-of-the-art performance across numerous large-scale benchmarks, outperforming both classical and learning-based 2D-3D lifting methods.
\end{abstract}

\begin{comment}
The recovery of 3D shape and pose from 2D landmarks stemming from a large ensemble of images can be viewed as a non-rigid structure from motion (NRSfM) problem.
Classical NR\SfM approaches, however, are problematic as they rely on heuristic priors on the 3D structure (e.g. low rank) but do not scale well to large datasets. Learning-based methods are showing the potential to reconstruct a much broader set of 3D structures -- dramatically expanding the importance of NRSfM to atemporal unsupervised 2D to 3D lifting. Hitherto, these learning approaches have not been able to effectively model perspective cameras or handle missing/occluded points -- limiting their applicability to in-the-wild datasets.
In this paper, we present a generalized strategy for improving learning-based NRSfM methods to tackle the above issues. Our approach, Deep NRSfM++, achieves state-of-the-art performance across numerous large-scale benchmarks, outperforming both classical and recent 2D-3D lifting methods.
\end{comment}

%% file: sections/1-intro.tex
\section{Introduction}

Non-rigid Structure from Motion (NR\SfM) aims to reconstruct the 3D structure of a deforming object from 2D keypoint correspondences observed from multiple views~\cite{akhter2009nonrigid,dai2014simple,kumar2016multi,bregler2000recovering,kumar2019jumping,kong2016prior}.
While the object deformation has classically been assumed to occur in time, the vision community has increasingly drawn attention to \emph{atemporal} applications -- commonly known as unsupervised 2D-3D lifting.
Notable examples include (i) Structure from Category (\SfC)~\cite{kong2016structure,agudo2016recovering,Agudo_2018_CVPR,gao2016symmetric}, where the deformation exists as the shape variation within a category-specific object dataset, and (ii) unsupervised human pose estimation~\cite{chen_drover,posegan,Wang_2019_ICCV}, which aims to recover the 3D human structure from 2D keypoint annotations.
% where a single-view 2D-to-3D pose lifting estimator is learned from 2D keypoint annotations.}
More recently, learning-based methods~\cite{c3dpo,ck19,cha2019unsupervised} have shown impressive results for solving atemporal NR\SfM~problems.
% {\chc what are the exact differences? not clear}
These models are advantageous because (i) they are scalable to large datasets and (ii) they allow fast feed-forward predictions once trained, without the need of rerunning costly optimization procedures for new samples during inference.
% not requiring the running of a costly optimization procedures for new samples once a model has been learned for an object category.

In this regard Deep NR\SfM~\cite{ck19} is of particular interest.
% Of particular interest in this regard is the recent Deep NR\SfM~\cite{ck19} method of Kong \& Lucey.
% Unlike other recent deep methods -- most notably C3PDO~\cite{c3dpo} -- Deep NR\SfM~is a factorization method at its heart.
Deep NR\SfM~at its heart is a factorization method that learns a series of dictionaries end-to-end for the joint recovery of 3D shapes and camera poses.
Unlike classical NR\SfM~approaches, it incorporates hierarchical block sparsity as the shape prior, which was shown to be more expressible than the popular low-rank assumption~\cite{dai2014simple,kumar2020non,bregler2000recovering,akhter2009defense,fragkiadaki2014grouping} and more robust than single-level block sparsity~\cite{kong2016structure,kong2016prior}.
A key concept in Deep NR\SfM~is its imposition of this prior in the \emph{architectural design} of the network.
This allows one to solve the hierarchical block-sparse dictionary learning problem by solely optimizing the reprojection objective end-to-end using deep learning as the machinery.
This also makes such structured bias in the network architecture to be interpretable as the optimization procedure used for estimating block-sparse codes.
% Deep NR\SfM~solves the hierarchical block-sparse dictionary learning problem by optimizing the reprojection objective end-to-end using deep learning as the machinery, where the structured inductive biases in the network architecture reflect the optimization procedure used for estimating block-sparse codes.
Deep NR\SfM~has shown superior performance on \SfC and temporally unordered motion capture datasets, outperforming comparable NR\SfM~methods~\cite{gotardo2011kernel,hamsici2012learning,lee2016consensus,Agudo_2018_CVPR} by an order of magnitude.

%Many of these works originates from unsupervised pose estimation, where they have a black-box pose estimator learned with a combination of reprojection error and prior constraints (\eg low rank~\cite{cha2019unsupervised}, consensus~\cite{c3dpo}, and adversarial discriminator~\cite{chen_drover}). Although achieving success in a number of \SfC and pose estimation tasks, limited by lack of interpretability in their models, they miss the opportunity of utilizing the geometric constraints from the architecture level.

%Although achieving success in a number of \SfC and pose estimation tasks, limited by lack of geometric interpretability in their models, it is mathematically less intuitive to extend their method for data with strong perspective effect~\cite{c3dpo} or missing keypoint annotations due to heavy occlusions~\cite{chen_drover,}, which are real concerns in practise.

Despite the recent advances on NR\SfM, however, two common drawbacks to state-of-the-art NR\SfM methods~\cite{kumar2016multi,kumar2020non,gotardo2011kernel,hamsici2012learning,gotardo2011non} still remain: (i) the assumption of an \emph{orthogonal camera model} and (ii) the need for cumbersome post-processing (\eg low-rank matrix completion) to handle \emph{missing data} (keypoints).
This makes them impractical for most datasets collected in the wild, where images can exhibit strong perspective effects and structured missing keypoint annotations due to self/out-of-view occlusions.
Although perspective effects have been studied in physics-based NR\SfM~\cite{parashar2018self,kumar2017monocular,vicente2012soft,chhatkuli2017inextensible} where priors such as isometric deformation are enforced, they rarely hold for generic atemporal shapes (\eg object categories in \SfC) under consideration in this paper. 
\begin{comment}
A drawback to the approach as well as most state-of-the-art NR\SfM~methods~\cite{kumar2016multi,kumar2020non,gotardo2011kernel,hamsici2012learning,gotardo2011non}, however, is the assumption of an orthogonal camera model, and the need for cumbersome post-processing (\ie low rank matrix completion) to handle missing data. This makes it impractical for most datasets collected in the wild, where images can: (i) exhibit strong perspective effects; and (ii) structured missing keypoint annotations due to self/out-of-view occlusions. Perspective effects have been previously studied in NR\SfM literature -- most notably in physics-based NR\SfM~\cite{parashar2018self,kumar2017monocular,vicente2012soft,chhatkuli2017inextensible} where priors such as isometric deformation are often enforced. Unfortunately, such priors rarely hold for general atemporal shapes (\eg object categories in SfC) under consideration in this paper. 
\end{comment}
The application of a perspective camera model to factorization methods, on the other hand, is nontrivial.
Specifically, the bilinear relationship of 2D observations to the factorized 3D shape and camera matrices in~\eqref{eq:w=sp} breaks down; as a consequence, the na\"ive block sparsity prior would no longer mathematically hold.

In this paper, we address the above issues of recent learning-based NR\SfM methods and provide a solution to incorporate perspective cameras and handle missing data at a theoretical angle.
We focus on Deep NR\SfM~\cite{ck19} and show that the bilinear factorization relationship in~\eqref{eq:w=sp} can be preserved by centering the common reference frame to the object center.
This seemingly simple step, which draws inspiration from Procrustes analysis~\cite{gower1975generalized,bartoli2013stratified,lee2013procrustean} and previously applied in Perspective-n-Point (PnP) problem~\cite{zheng2013revisiting}, circumvents the need for explicitly modeling translation and projective depths in NR\SfM.
% Furthermore, we show that this object-centric recasting of the NR\SfM~factorization problem can be made effective for handling missing points. 
The resulting bilinear factorization induces a novel 2D-3D lifting network architecture which is adaptive to camera models as well as the visiblity of input 2D points.
% Finally, we demonstrate how this generalization can be explicitly incorporated into~\cite{ck19}, which we refer to as \textbf{Deep NR\SfMpp}, resulting in state-of-the-art results across a myriad of benchmarks. 
We refer to our approach as \textbf{Deep NR\SfMpp}, which achieves state-of-the-art results across a myriad of benchmarks. 
In addition, we demonstrate significant improvements of Deep NR\SfMpp~in reconstruction accuracy by correcting an inherent caveat of Deep NR\SfM, which made unwarranted relaxations of the block sparsity prior that deviates from the true block-sparse coding objective.

%We show that missing data and perspective projection can be accounted for by adaptively normalizing both the input 2D keypoints and the shape dictionary; in addition, explicit estimation of the camera's translational component can be circumvented by fully taking advantage of the object-centric nature of the problem.
%These reformulations lead to a unified framework under both strong and weak perspective camera models, capable of handling missing data.

Our contributions are summarized as below:
\begin{itemize}
    \item We derive a bilinear factorization formulation to incorporate perspective projection and handle missing data in learning-based NR\SfM~methods.
    % The innovation of our approach is two-fold: (i) We avoid explicitly estimating the camera's translational component by fully taking advantage of the object-centric nature of the problem. (ii) Missing data and perspective projection are accounted in the form of adaptively normalizing both the 2D input and the shape dictionary. These reformulations lead to a unified framework under both perspective and weak perspective camera, capable of handling missing data. 
    \item We propose a solution at the architectural level that keeps a closer mathematical proximity to the hierarchical block-sparse coding objective in NR\SfM.
    % \item We show that it is empirically beneficial to keep even closer proximity to the hierarchical block sparse coding objective.
    % \item We propose a novel method to handle both missing data and strong perspective camera models via mathematical reformulations, avoiding the need to explicitly estimate translational components while being more compatible to block sparse coding in NR\SfM.
    % variation of Strum \& Triggs~\cite{sturm1996factorization} that is more compatible to block sparse coding.
    \item We outperform numerous classical and learning-based NR\SfM methods and achieve state-of-the-art performance across multiple benchmarks, showing its effectiveness in handling large amounts of missing data under both weak and strong perspective camera models.
\end{itemize}

%\noindent We note that in this paper, we are considering approaches to NR\SfM~that make no assumptions about the temporal relationship between images and more generally applicable to datasets that are disjoint in both space and time. 

\begin{comment}
(i) We show that it is empirically beneficial to keep even closer proximity to the hierarchical block sparse coding objective.\\
(ii) We offer a new formulation to handle missing data and perspective projection under the context of shape from category. By fully utilizing the assumption of being object centric, we cancels out the need for explicitly estimating translation. And in terms of dealing with perspective projection, we propose a new variation of Strum \& Triggs~\cite{sturm1996factorization} that is more compatible to block sparse coding.\\
(iii) We demonstrate SOTA performance across multiple benchmarks when comparing against NR\SfM~and deep learning methods. From those empirical results, our method show its effectiveness in handling high percentage of missing data under both weak perspective and perspecive camera models.
\end{comment}
%(iii) From the network architecture perspective, our work demonstrate a new way of handeling 2D input with missing data or perspective effect. Instead of either feeding in binary visibility masks or expect the network to implicitly learn the underline geometric relation, we propose to shift and scale the network weights to account for the variations in the input data.

%% file: sections/2-relatedwork.tex
\section{Related Work}
\vspace{6px}
\noindent \textbf{Non-rigid structure from motion. }
NR\SfM concerns the problem of reconstructing 3D shapes from 2D point correspondences from multiple images, \emph{without} the assumption of the 3D shape being rigid. For a shape of $P$ points under orthogonal projection, \emph{atemporal} NR\SfM is typically framed as factorizing the 2D measurements $\W\in\mathbb{R}^{P\times2}$ as the product of a 3D shape matrix $\S\in\mathbb{R}^{P\times3}$ and camera rotation matrix $\mathbf{P}\in\mathbb{R}^{3\times2}$:
\begin{equation}
    \W = \S\mathbf{P} \;\;\;\;\text{subject to}\;\; \mathbf{P}^{\top}\mathbf{P}=\I_2 \;,
    \label{eq:w=sp}
\end{equation}
% \begin{equation}
% \W = \begin{pmatrix}\vdots & \vdots \\ u_i & v_i \\ \vdots & \vdots \end{pmatrix},~~
% \S = \begin{pmatrix} \vdots & \vdots & \vdots \\ x_i & y_i & z_i \\ \vdots & \vdots &\vdots \end{pmatrix}, \mathbf{P}^{\top}\mathbf{P} = \I_2
% \end{equation}
where $\I$ is the identity matrix, and the $i$-th row of $\W$ and $\S$ corresponds respectively to the image coordinates $(u_i, v_i)$ and world coordinates $(x_i,y_i,z_i)$ of the $i$-th point.

This factorization problem is ill-posed by nature; in order to resolve the ambiguities in solutions, additional priors are necessary to guarantee the uniqueness of the solution.
 % are enforced {\red upon the stack of shape matrices under multiple views \chc what does this mean?}, and also on the trajectories if temporal information is available.
These priors include the assumption of shape/trajectory matrices being (i) low-rank~\cite{dai2014simple,bregler2000recovering,akhter2009defense,fragkiadaki2014grouping,kumar2020non}, (ii) being compressible~\cite{kong2016structure,kong2016prior,zhou2016sparseness}, or (iii) lying in a union of subspaces~\cite{kumar2016multi,zhu2014complex,Agudo_2018_CVPR}.
Although classical NR\SfM methods incorporating such priors are mathematically well interpreted, they encounter limitations in large-scale datasets (\eg with a large number of frames).
The low-rank assumption becomes infeasible when the data exhibits complex shapes variations.
 % {\red and the number of points is much smaller than the number of frames}.
Union-of-subspaces NR\SfM methods have difficulty clustering shape deformations and estimating affinity matrices effectively.
 % when the number of frames is large.
The sparsity prior allows more powerful modeling of shape variations with large number of subspaces but also suffers from sensitivity to noise.
% Low-rank assumption is infeasible when the data has complex variations and the number of points is much smaller than the number of frames; Union-of-subspaces NR\SfM has difficulties in how to effectively cluster shape deformations solely from 2D inputs, and how to estimate affinity matrix when the number of frames is large; Sparsity prior enables more power to model complex shape variations with large number of subspaces. But because there are many possible subspaces to choose from, it is sensitive to noise.

\vspace{6px}
\noindent \textbf{Perspective projection. }
Most factorization-based NR\SfM research assumes an orthogonal camera model, but this assumption often breaks down for real-world data, where strong perspective effects may exhibit (\eg when objects are sufficiently close to the camera).
Modeling perspective projection thus become necessary for more accurate 3D reconstruction.
% But in the real-world data that has objects close to the camera, modeling perspective projection is necessary for accurate reconstruction.
Sturm \& Triggs~\cite{sturm1996factorization} formulate the rigid \SfM problem under perspective camera as
\begin{equation}
    \text{diag}(z_1, \dots, z_P) [\W \;\; \mathbf{1}_P] = [\S \;\; \mathbf{1}_P] \tilde{\mathbf{P}} \;,
    % \text{diag}(z_1, z_2, \dots z_P) \begin{bmatrix} \W & \mathbf{1}_P \end{bmatrix} = \begin{bmatrix}\S& \mathbf{1}_P\end{bmatrix} \tilde{\mathbf{P}}.
    \label{eq:zw=sp}
\end{equation}
where $z_i$ is the projective depth of the $i$-th point and the camera matrix $\tilde{\mathbf{P}}\in\mathbb{R}^{4\times3}$ additionally models translation.
This is similar to~\eqref{eq:w=sp} except that the 2D measurements $\W$ is now multiplied by the unknown depth.
Since the problem is nonlinear, iterative optimization on the reprojection error is adopted to adjust the projective depth while adding constraints on the rank~\cite{sturm1996factorization,oliensis2007iterative,christy1996euclidean,ueshiba1998factorization,heyden1999iterative}.
 % Sturm \& Triggs~\cite{sturm1996factorization} and others~\cite{oliensis2007iterative,christy1996euclidean,ueshiba1998factorization,heyden1999iterative} iteratively adjust the projective depth using the rank constraint while optimizing the reprojection error.

This formulation was later extended to solve NR\SfM.
Xiao \& Kanade~\cite{xiao2005uncalibrated} developed a two-step factorization algorithm by recovering the projective depths with subspace constraints and then solving the factorization with orthogonal NR\SfM. %factorize $\text{diag}(z_1, z_2, \dots z_P)\W$ by weak perspective NR\SfM.  
Wang~\etal~\cite{wang2007structure} proposed to update solutions by gradually increasing the perspectiveness of the camera and recursively refining the projective depth.
Hartley \& Vidal~\cite{hartley2008perspective} derived a closed-form linear solution requiring an initial estimation of a multi-focal tensor, which was reported sensitive to noise.
% Other advances in rigid \SfM (\eg \cite{dai2013projective,magerand2017practical}) are potentially helpful to factorization-based NR\SfM, but are largely restricted to the low-rank methods.
Instead of directly solving the problem in~\eqref{eq:zw=sp}, we simplify the problem by fully utilizing the object-centric constraint, which has been previously used to remove projective depth and translaton in PnP~\cite{zheng2013revisiting}. This results in a bilinear reformulation which is more compatible to factorization-based methods.
% -- we cancels out the translation inside the camera projection matrix, and removes the need to estimate the projective depth as a separate variable v.s. the shape matrix. This leads to a novel formulation that is compatible to block sparse coding.
%a linear equation with unkowns only on one side of the equation, and thus making the problem more convenient to deal with.

% We note that although perspective effects have also been studied in physics-based NR\SfM~\cite{parashar2018self,kumar2017monocular,vicente2012soft,chhatkuli2017inextensible} where priors such as isometric deformation are enforced, they rarely hold for generic atemporal shapes (\eg object categories in \SfC) under consideration in this paper. 

\vspace{6px}
\noindent \textbf{Missing data. }
Missing (annotated) point correspondences are inevitable in real-world data due to self/out-of-view occlusions of objects.
Handling missing data is a nontrivial task not only because the input data is incomplete, but also because the true object center becomes unknown, making na\"ive centering of 2D data an ineffective strategy for compensating translation.
Previous works employ a strategy of either (i) introducing a visibility mask to exclude missing points from the objective function~\cite{del2007non,gotardo2011kernel,kong2016structure,Wang2008RotationCP,ck19}, (ii) recovering the missing 2D points with matrix completion algorithms prior running NR\SfM~\cite{dai2014simple,lee2016consensus,hamsici2012learning,kumar2020non}, or (iii) treating missing points as unknowns and updating them in an iterative fashion~\cite{paladini2012optimal}.
We follow the first strategy and introduce a visibility mask, but with an additional object-centric constraint~\cite{lee2013procrustean}.
Our key difference to Lee~\etal~\cite{lee2013procrustean} is that we (i) derive a bilinear form with dictionaries adaptively normalized to the data; (ii) extend to handle perspective projection which is missing in~\cite{lee2013procrustean}.

% {\red In this work, we follow the first strategy, and derives a novel approach under the framework of block sparse coding. \ch this sounds weak, need another sentence stating our actual contribution in this regard}
% which leads to an implementation as a feedforward neural network.
% This is similar to the approach by Kong \& Lucey~\cite{ck19} except the way how we handle the translation -- instead of either pretending the translation to be zero or treating it as unknown, we cancel the translation by assuming that it equals to the object center position. This object-centric view simplifies the problem, and leads to improved empirical results.

\vspace{6px}
% <<<<<<< HEAD
% \noindent \textbf{Deep learning methods for NR\SfM. }
% {\ch
% Unsupervised 2D-3D lifting, which is equivalent to \emph{atemporal} NR\SfM, is recently approached by training neural networks to predict the depth coordinate of each input 2D point.
% }
% Such lifting networks are trained by minimizing the 2D reprojection error, but as with classical NR\SfM problems, this is inherently an ill-posed problem that requires priors or constraints on the loss in certain forms.
% % However 2D reprojection error alone is obviously not enough to constrain the problem, and thus other constraints are needed.
% {\ch
% One popular prior is the use of Generative Adversarial Networks (GANs)~\cite{goodfellow2014generative} to enforce realism of 2D reprojections across novel viewpoints~\cite{chen_drover,repnet,drover,posegan}.
% The recently proposed C3DPO~\cite{c3dpo}, on the other hand, instead {\red enforces additional self-consistency on the predicted canonicalization of the same 3D shapes}.
% }
%  % and various forms of self-consistency loss~\cite{c3dpo,chen_drover,repnet,drover,posegan}, are used to assist training.
% These methods, however, learn to tolerate both missing data and perspective effects from data, making the designed models geometrically uninterpretable.
% =======
\noindent \textbf{Unsupervised 2D-3D lifting}
% {\ch
The problem of unsupervised 2D-3D lifting, which is equivalent to \emph{atemporal} NR\SfM, is recently approached by training neural networks to predict the third (depth) coordinate of each input 2D point.
% }
These networks are trained by minimizing the 2D reprojection error, but as with classical NR\SfM, this is inherently an ill-posed problem that requires priors or constraints in certain forms.
% However 2D reprojection error alone is obviously not enough to constrain the problem, and thus other constraints are needed.
One popular prior is the use of Generative Adversarial Networks (GANs)~\cite{goodfellow2014generative} to enforce realism of 2D reprojections across novel viewpoints~\cite{chen_drover,repnet,drover,posegan}.
Low-rank priors used in classical NR\SfM was also explored as a loss function~\cite{cha2019unsupervised}.
The recently proposed C3DPO~\cite{c3dpo} instead enforces self-consistency on the predicted canonicalization of the same 3D shapes.
 % and various forms of self-consistency loss~\cite{c3dpo,chen_drover,repnet,drover,posegan}, are used to assist training.
These methods use black-boxed network architectures without geometric interpretability, and thus their robustness to missing data and perspective effects depends solely on the variations within the data, making learning inefficient.
In addition, most of these methods~\cite{chen_drover,posegan,cha2019unsupervised,drover,repnet} are incapable of handling missing data. 
%~\cite{chen_drover,posegan,cha2019unsupervised,drover,repnet}.
% These methods are developed mostly from the machine learning point of view, and as a consequence, lack geometric interpretability especially when dealing with both missing data and perspective projection.
We mathematically derive a general framework which is applicable for both orthogonal and perspective cameras, robust to large portion of missing data, and interpretable as solving hierarchical block-sparse dictionary coding.

%% file: sections/3-method.tex
\section{Generalized Bilinear Factorization} 
\label{sec:bilinear}

% =============================================================

% \subsection{\ch Generalized Bilinear Formulation}
% \label{sec:generalize}

A wide range of 2D-3D lifting methods assume a linear model for the 3D shapes to be reconstructed, \ie at canonical coordinates, the vectorization of 3D shape $\S$ in~\eqref{eq:w=sp}, denoted as $\s = \text{vec}(\S) \in \Real^{3P}$, can be written as $\s = \D\bvarphi$, where $\D\in \Real^{3P\times K}$ is the shape dictionary with $K$ basis and $\bvarphi\in\Real^K$ is the code vector.
Equivalently, this linear model could be writen as $\S = \D^\sharp (\bvarphi \otimes \I_3)$, where $\D^\sharp$ is a $P\times3K$ reshape of $\D$ and $\otimes$ denotes Kronecker product.
Applying the camera extrinsics (\ie rotation $\R\in\SO(3)$ and translation $\t\in\Real^3$) gives the 3D reconstruction in the camera coordinates, written as
% A wide range of 2D-3D lifting methods assume a linear model for the 3D shapes to be reconstructed, \ie at canonical coordinates, 3D shape vector $\s = \D\bvarphi $, in which $\s$ is the vectorization of $\S$ in (\ref{eq:w=sp}), $\D\in \Real^{3P\times K}$ is the shape dictionary with $K$ basis, and $\bvarphi\in\Real^K$ is the code vector.
% Equivalently, this linear model could be writen as $\S = \D^\sharp (\bvarphi \otimes \I_3)$, where $\D^\sharp$ is a $P\times3K$ reshape of $\D$, and $\otimes$ denotes Kronecker product. Applying the camera extrinsics (\ie rotation $\R\in\text{so}(3)$, translation $\t\in\Real^3$) results in the 3D reconstruction at camera coordinates:
\begin{equation}
    \S_\text{cam} = \D^\sharp (\bvarphi \otimes \R) + \1_P \t^\top \;,
\end{equation}
where $\1$ is the one vector.
Under unsupervised settings, $\D$, $\bvarphi$, $\R$, and $\t$ are all unknowns and are usually solved under simplified assumptions, \ie complete input 2D points under orthogonal camera projection.
In addition, the translational component $\t$ could be removed if the input 2D points were pre-centered under the implicit object-centric assumption that the origin of the canonical coordinates is placed at the object center.
% \cy{Under unsupervised settings, $\D$, $\bvarphi$, $\R$, $\t$ are the unkowns, and usually solved under simplified assumptions, \ie the camera projection is orthogonal and the input 2D points are complete with no occlusion. In addition, with the \emph{object centric} assumption induced from the linear model, which implicitly assumes that the origin of canonical coordinates are placed at the center of an object, the translational component $\t$ can thus be removed if the input 2D points are pre-centered.
This leads to a bilinear factorization problem which classical NR\SfM methods~\cite{ck19,dai2014simple,kumar2020non} employ:
\begin{equation}
    \W = \D^\sharp \bPsi_\xy \;\;\;\; \text{s.t.}\;\; \bPsi=\bvarphi\otimes\R \;\;\text{and}\;\; \bvarphi \in \C \;,
    \label{eq:w=dpsi}
\end{equation}
where $\bPsi_\xy\in\Real^{3\times2}$ denotes the first two columns of $\bPsi$, $\bPsi \in \Real^{3K\times3}$ is the block code (as it is a Kronecker product), and $\C$ denotes the prior constraints applied on the code $\bvarphi$, \eg low rank~\cite{dai2014simple,kumar2017monocular} or (hierarchical) sparsity~\cite{kong2016structure,kong2016prior,ck19}.

In practical settings, however, the presence of either \emph{missing data} (from occlusions) or \emph{perspective projection} would break this bilinear form.
With missing data in the 2D points $\W$, the translation $\t$ does not vanish if $\W$ is simply pre-centered by the average of visible points.
Moreover, directly applying Strum \& Triggs~\cite{sturm1996factorization} under perspective projection would introduce unknown projective depth $z_i$, resulting in a non-bilinear form of
% Second, under perspective projection, with the introduction of unknown projective depth, directly applying Strum \& Triggs~\cite{sturm1996factorization} results in a non-bilinear form:
\begin{equation}
    \text{diag}(z_1, \dots, z_P) \W = \D^\sharp \bPsi_\xy + \1_P\t_\xy^\top \;,
    \label{eq:strum_triggs_nrsfm}
\end{equation}
which prevents direct application of NR\SfM algorithms designed for the orthogonal case.
% {\ch
Iterative methods which alternates between solving orthogonal NR\SfM and optimizing $z_p$~\cite{wang2007structure} can be applied, but it is cumbersome and prone to poor local minima.
% {\chc (related work removed)}
On the other hand, similar issue is also encountered in PnP problems, where the projective depths make the problem nonlinear. A simple and effective solution for PnP is to turn the implicit object-centric assumption into an \emph{explicit} constraint, \ie: 
% One of the solutions is simple and effective, \ie turn the previously implicit object-centric assumption to be an explict hard constraint, which induces a simple equation:
\begin{equation}
    \t = \frac{1}{P}{\S_\text{cam}}^{\top}\1_P \;,
    \label{eq:t=mean}
\end{equation}
The introduction of this simple equation allows the projective depth and translation to be eliminated for PnP~\cite{zheng2013revisiting}. Similar tricks also apply to NR\SfM to remove $z_i$ and $\t$ in~\eqref{eq:strum_triggs_nrsfm}, and naturally extends to handle missing data. Hence, we derive a generalized bilinear factorization, which considers both orthogonal and perspective projections, as well as missing data:
\begin{equation}
    \M\tW = \M\tD \tbPsi \;\;\;\; \text{s.t.}\;\; \mathbf{R}\in \SO(3) \;\;\text{and}\;\; \bvarphi\in\C \;,
    \label{eq:general_bilinear}
\end{equation}
where $\M=\text{diag}(m_1,\cdots,m_P)$ is the input binary diagonal matrix indicating visibility of points; and $\tW$, $\tD$, $\tbPsi$ are the transformed forms of $\W$, $\D$, $\bPsi$, whose formulation under different settings are summarized in Table~\ref{tab:math_summary}. Mathematical derivations are provided in Sec.~\ref{sec:persp} for perspective camera and Sec.~\ref{sec:missing_data} for missing data.

\begin{table*}[t!]
\small
    \centering
    \def\arraystretch{1.3}
    \begin{tabular}{|c|c|c|c|c|c|c|}
    \hline
    & \multicolumn{2}{c|}{$\tW$} & \multicolumn{2}{c|}{$\tD$} & \multicolumn{2}{c|}{$\tbPsi$} \\ \cline{2-7}
    & formulation & shape & formulation & shape & formulation & shape \\ \hline
    &&&&&&\\[-12pt]
    orthogonal & $\W -  \frac{1}{\tP} \1_P\1_P^{\top} \M\W$ &  $\Real^{P\times 2}$ & $\D^\sharp + \frac{1}{\tP}\1_P\1_P^{\top} (\I_P - \M)\D^\sharp$ &  $\Real^{P\times 3K}$ & $\bvarphi\otimes\R_\xy$ & $\Real^{3K\times 2}$ \\ 
    &&&&&&\\[-12pt] \hline
    perspective & \eqref{eq:eq_persp} &  $\Real^{2P\times 1}$ & \eqref{eq:eq_persp} &  $\Real^{2P\times 9K}$ & $\text{vec}(\bvarphi\otimes\R)$ & $\Real^{9K\times 1}$ \\
    \hline
    \end{tabular}
    \vspace{-4pt}
    \caption{Summary of the formulations of matrices $\tW$, $\tD$, and $\tbPsi$ under orthogonal and perspective cameras.}
    \label{tab:math_summary}
    % \vspace{-8pt}
\end{table*}

% =============================================================

\subsection{Perspective Camera}
\label{sec:persp}
We first consider the case where all points are visible.
Let $(x'_i, y'_i, z'_i)$ be the 3D coordinates of the $i$-th point in $\S\R$.
Since $\S\R = \D^\sharp \bPsi$, we can also express $(x'_i, y'_i, z'_i)$ as
\begin{equation}
    x'_i = \d^{\top}_i \bpsi_x,~~~y'_i = \d^{\top}_i \bpsi_y,~~~z'_i = \d^{\top}_i \bpsi_z \;,
\label{eq:x=dpsi}
\end{equation}
where $\d_i^T$ is the $i$-th row of $\D^\sharp$ and $\bpsi_x$, $\bpsi_y$, and $\bpsi_z$ are the three columns in $\bPsi$ corresponding to the $i$-th point.
The 3D point $(x'_i, y'_i, z'_i)$, its 3D translation $(t_x, t_y, t_z)$, and its 2D projection $(u_i,v_i)$ on the unit focal plane are related via
% Assuming that camera intrinsics are given, or equivalently $\mathbf{K} = \I_3$,  we have the linear relationships:
% \begin{equation}
%     \begin{aligned}
%     u_i = \frac{x'_i + t_x}{z'_i + t_z},~~~
%     v_i = \frac{y'_i + t_y}{z'_i + t_z}\\
%     \end{aligned}
% \end{equation}
% We can make this equation linear by multiplying $z'_i+t_z$ on both sides:
\begin{equation}
    x'_i + t_x  = u_i(z'_i + t_z) \;\;\;\text{and}\;\;\;
    y'_i + t_y  = v_i(z'_i + t_z) \;,
    \label{eq:basic_eq1}
\end{equation}
which states that the product of the depth and 2D coordinates is equivalent to the x-y coordinates in 3D. From the object-centric constraint in~\eqref{eq:t=mean}, the translation can be expressed as the mean of back-projection of the 2D points as
\begin{equation}
    t_x  = \frac{1}{P}\sum_{i=1}^P u_i(z'_i+t_z) \;\;\text{and}\;\;
    t_y  = \frac{1}{P}\sum_{i=1}^P v_i(z'_i+t_z) .
    \label{eq:translation}
\end{equation}
% Substituting~\eqref{eq:translation} and~\eqref{eq:x=dpsi} into~\eqref{eq:basic_eq1} and rearranging, we obtain the following linear relationships in matrix form (written compactly as $\tW = \tD \tbPsi$):
Substituting~\eqref{eq:translation} and~\eqref{eq:x=dpsi} into~\eqref{eq:basic_eq1} and rearranging, we obtain the compact bilinear relationship of $\tW = \tD \tbPsi$, written as
\begin{equation}
    \resizebox{0.96\hsize}{!}{$
    % \scriptsize
    \setlength\arraycolsep{1pt}
    \everymath{\displaystyle}
    \underbrace{
    \begin{pmatrix}
    \vdots\\
    (u_i - \frac{1}{P}\sum_{j=1}^P u_j) t_z\\
    (v_i - \frac{1}{P}\sum_{j=1}^P v_j) t_z\\
    \vdots
    \end{pmatrix}
    }_{\textstyle \tW \text{: normalized 2D projection}}
    =  \underbrace{\begin{pmatrix}
      \vdots & \vdots & \vdots \\
        \d_i^{\top} & 0 & -u_i\d_i^{\top}+\frac{1}{P}\sum_{j=1}^Pu_j\d_j^{\top}\\
        0 & \d_i^{\top} & -v_i\d_i^{\top}+\frac{1}{P}\sum_{j=1}^P v_j\d_j^{\top}\\
        \vdots & \vdots & \vdots
    \end{pmatrix}}_{\textstyle \tD \text{: normalized dictionary}}
    \underbrace{
     \begin{pmatrix}\bpsi_x \\
    \bpsi_y\\
    \bpsi_z
    \end{pmatrix}
    }_{\tbPsi}
    $} \;,
    \label{eq:persp_basic}
\end{equation}
% \begin{align}
%     \tW &= \begin{pmatrix}
%     \vdots\\
%     (u_i - \frac{1}{P}\sum_{j=1}^P u_j) t_z\\
%     (v_i - \frac{1}{P}\sum_{j=1}^P v_j) t_z\\
%     \vdots
%     \end{pmatrix} \;, \\
%     \tD &= \begin{pmatrix}
%       \vdots & \vdots & \vdots \\
%         \d_i^{\top} & 0 & -u_i\d_i^{\top}+\frac{1}{P}\sum_{j=1}^Pu_j\d_j^{\top}\\
%         0 & \d_i^{\top} & -v_i\d_i^{\top}+\frac{1}{P}\sum_{j=1}^P v_j\d_j^{\top}\\
%         \vdots & \vdots & \vdots
%     \end{pmatrix} \;.
%     \label{eq:persp_basic2}
% \end{align}
where $\tW$ is computed from $\W$ via zero-centering by the mean and rescaling by $t_z$, the depth of the object center to the camera.
Here, $t_z$ is a scalar that normalizes the 2D input and controls the scale of the 3D reconstruction, which is similar to the weak perspective case.
$\tbPsi$ becomes a vectorization of $\bPsi$ where the columns are simply concatenated.

\vspace{6px}
\noindent \textbf{Shape scale $t_z$. }
As shown in \eqref{eq:persp_basic}, $t_z$ serves the purpose of normalizing the 2D inputs and consequently sets the scale of the shape reconstruction.
We note that unlike rigid \SfM or trajectory reconstruction, where scale constancy between frames (\eg object staticity or temporal smoothness within the same dataset) is important, the purpose of ``estimating'' scale is fundamentally inapplicable to this work.
Under the atemporal NR\SfM settings, where we deal with non-rigid objects only in the shape space, it is sufficient to obtain a scaled 3D reconstruction, as was in the weak perspective case. This in turn allows us to determine an arbitrary scale $t_z$ for each sample as a preprocessing step to facilitate training.
We leave details of determining $t_z$ in practice to Sec.~\ref{sec:exp}.

\vspace{6px}
\subsection{Handling Missing Data} 
\label{sec:missing_data}
% \vspace{6px}
\noindent \textbf{Perspective camera. }
When all the 2D keypoint locations might not be fully available, it becomes insufficient to compute $\t$ from~\eqref{eq:translation}.
To resolve this, we propose to replace the occluded keypoints with the ones directly from $\S_\text{cam}$, \ie%, \ie the translation is expressed using the average of occluded 3D points as well as 2D visible points multiplied by their projective depth}:
\begin{align}
    t_x &= \frac{1}{P}\sum_{i=1}^P \underbrace{ m_i u_i (z'_i + t_z)}_{\text{visible points}} + \underbrace{(1-m_i)(x'_i + t_x)}_{\text{occluded points}} \;,
    \label{eq:visibility}
\end{align}
where $m_i$ indicates the visibility.
Rearranging~\eqref{eq:visibility}, we have
\begin{equation}
\resizebox{0.6\hsize}{!}{$
   t_x = \frac{ \sum_{i=1}^P m_i u_i (z'_i + t_z) + (1-m_i)x'_i } {\sum_{i=1}^P m_i}
   $}
\end{equation}
and similarly for $t_y$.
Substituting the new expressions of the translational components $t_x$ and $t_y$ into~\eqref{eq:basic_eq1}, we have
\begin{equation}
    \resizebox{0.96\hsize}{!}{$
    % \scriptsize
    \setlength\arraycolsep{1pt}
    \everymath{\displaystyle}
    \M \underbrace{ \begin{pmatrix}
    \vdots\\
    (u_i - \frac{\sum_{j=1}^P m_j u_j }{\sum_{j=1}^P m_j})t_z\\
    (v_i - \frac{\sum_{j=1}^P m_j v_j }{\sum_{j=1}^P m_j})t_z\\
    \vdots
    \end{pmatrix} }_{\textstyle \tW \text{: normalized 2D projection}} = \M 
    \underbrace{
     \begin{pmatrix}
        \vdots & \vdots & \vdots \\
        \td_i^{\top}  & 0 & -u_i\d_i^{\top}+\frac{\sum_{j=1}^P m_j u_j\d_j^{\top}}{\sum_{j=1}^P m_j}\\
        0 & \td_i^{\top} & -v_i\d_i^{\top}+\frac{\sum_{j=1}^P m_j v_j\d_j^{\top}}{\sum_{j=1}^P m_j}\\
        \vdots & \vdots & \vdots
    \end{pmatrix}
    }_{\textstyle \tD \text{: normalized dictionary}}
    \begin{pmatrix}\bpsi_x \\
    \bpsi_y\\
    \bpsi_z
    \end{pmatrix} 
    $} \;,
    \label{eq:eq_persp}
\end{equation}
where $\td_i = \d_i + \frac{\sum_{j=1}^P (1- m_j) \d_j}{\sum_{j=1}^P m_j} $.

\vspace{6px}
\noindent \textbf{Orthogonal camera. }
% An identical strategy of handling missing data can be applied to orthogonal cameras.
Handling missing data under orthogonal cameras can be derived using as identical strategy as
\begin{equation}
    \resizebox{0.86\hsize}{!}{$
    % \scriptsize
    \setlength\arraycolsep{1pt}
    \everymath{\displaystyle}
    \M \underbrace{ (\W - \frac{1}{\tP}\1_P\1_P^{\top} \M\W)}_{\tW\text{: normalized 2D projection}} = 
    \M \underbrace{(\D^\sharp + \frac{1}{\tP}\1_P\1_P^{\top} (\I_P - \M)\D^\sharp)}_{\tD \text{: normalized dictionary}}\bPsi
    $} \;,
    \label{eq:orth_occlu}
\end{equation}
where $\tP$ is the number of visible points.
We leave the derivations to the supplementary material for conciseness.
This solution aligns with the common practice employed for data normalization~\cite{ck19,c3dpo,lee2013procrustean}. The difference is that we offset the shape dictionary as well to align the statistics to the shifted 2D inputs. %For the special case where all points are visible, the expressions in~\eqref{eq:orth_occlu} degenerates back to~\eqref{eq:w=dpsi}. 

%% file: sections/4-deepnrsfm++.tex
\section{Deep NR\SfMpp}

\begin{figure*}[t!]
  \centering
  \includegraphics[width=0.95\linewidth]{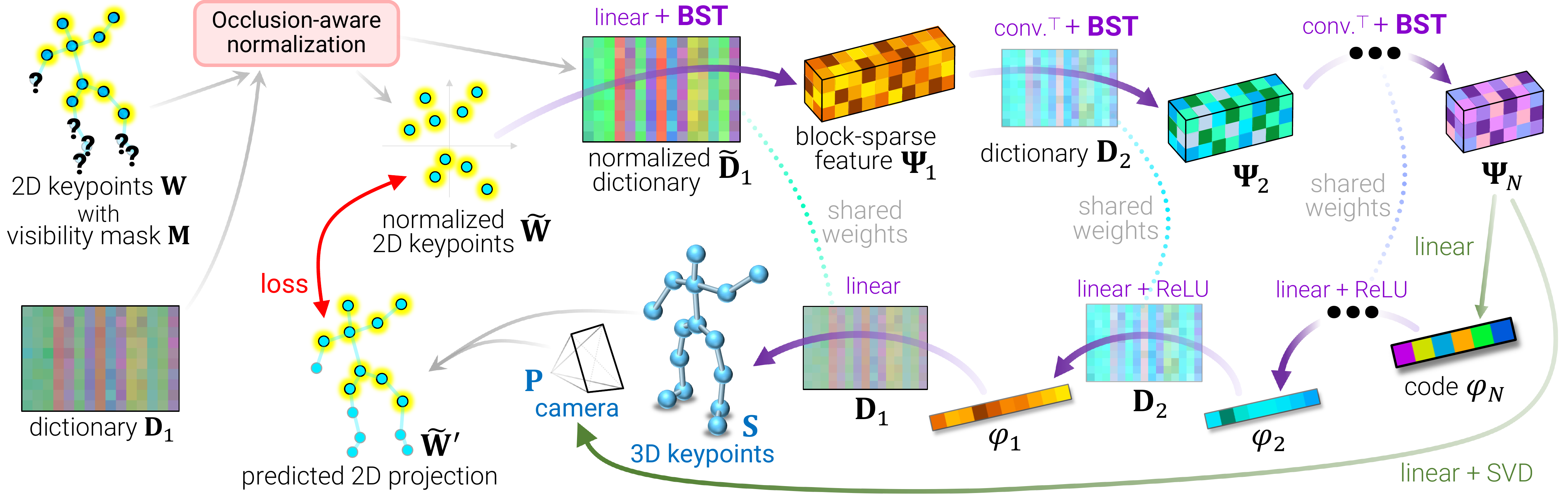}
  \caption{\textbf{Deep NR\SfMpp} is a general framework for learning hierarchical block-sparse dictionaries that operates under perspective camera models with the ability to handle missing data.
  The 2D keypoint input $\W$ and shape dictionary $\D_1$ are normalized according to the visibility mask $\M$ and camera model (Table~\ref{tab:math_summary}).
  The encoder-decoder network, derived from hierarchical block-sparse coding, takes the normalized 2D input $\tW$ and jointly predicts the camera matrix $\mathbf{P}$ and the 3D shape $\S$, further reprojected to 2D as $\tW'$.
  The training objective is to minimize the difference between $\tW'$ and $\tW$.
  }
  \label{fig:overview}
\end{figure*}

To solve the generalized bilinear factorization~\eqref{eq:general_bilinear} given $N$ tuples of $(\W^{(n)},\M^{(n)})$ as the dataset (with $n$ indexing the samples), two problems remain to address: (i) how to define heuristics $\C$, which is crucial to have an accurate solution, and (ii) how to formulate the optimization strategy. We choose to follow Deep NR\SfM~\cite{ck19}, which instead of using simple handcrafted priors (\eg low rank or sparsity), it imposes hierarchical sparsity constraint $\C_\Theta$ with learnable parameters $\Theta$ (see Sec.~\ref{sec:hbsc}). The learning strategy of Deep NR\SfM is then interpreted as solving a bilevel optimization problem:
\begin{equation}
\resizebox{0.88\hsize}{!}{$
\min_{\D,\Theta} \sum_{n=1}^N \min_{\substack{\bvarphi^{(n)}\in\mathcal{\C}_\Theta\\\R^{(n)}\in\text{SO}(3)}}\|\M^{(n)}\tilde{\W}^{(n)} - \M^{(n)}\tilde{\D} \tilde{\bPsi}^{(n)}\|_F,
$}
\label{eq:bilevel}
\end{equation}
where the lower level problems are to solve single frame 2D-3D lifting with given $\D$, $\Theta$; and the upper level problem is to find optimal $\D$,$\Theta$ for the whole dataset.

Descent method is employed to solve this bilevel problem. We first approximate the solver of the lower level problem as a feedforward network, \ie $f(\M, \W^{(n)}; \D,\Theta) \mapsto ({\R^*}^{(n)}_{\D,\Theta}, {\bvarphi^*}^{(n)}_{\D,\Theta})$. The architecture of the network (as illustrated in Fig.~\ref{fig:overview}) is induced from unrolling one iteration of Iterative Shrinkage and Thresholding Algorithm (ISTA)~\cite{beck2009fast,rozell2008sparse,hastie2005elements} (see Sec~\ref{sec:deepnrsfm++}). Then with ${\bPsi^*}^{(n)}_{\D,\Theta} = {\bvarphi^*}^{(n)}_{\D,\Theta} \otimes {\R^*}^{(n)}_{\D,\Theta}$, the original bilevel problem is reduced to a single level unconstrained problem, \ie
\begin{equation}
\resizebox{0.8\hsize}{!}{$
    \min_{\D, \Theta} \sum_{n=1}^N \|\M^{(n)}\tilde{\W}^{(n)} - \M^{(n)} \tilde{\D} {\tilde{\bPsi}^*}_{\D,\Theta}\|_F,$}
\end{equation}
which allows the use of solvers such as gradient descent. Finally, with $\D, \Theta$ learned, $f(\M,\W; \D,\Theta)$ is the 2D-3D lifting network applicable to unseen data.

\noindent\textbf{Note:} Due to the introduction of multiple levels of dictionaries and codes in Sec.~\ref{sec:hbsc}, we will abuse the notation of $\D$, $\bvarphi$, $\bPsi$ by adding subscript 1, \ie $\D_1$, $\bvarphi_1$, $\bPsi_1$ indicating that they are from the first level of the hierarchy.

% =============================================================

\subsection{Hierarchical Block-Sparse Coding (HBSC)}
\label{sec:hbsc}

Assuming the canonical 3D shapes are compressible via multi-layer sparse coding, the shape code $\bvarphi_1$ is constrained by $\C_\Theta$ as:
% \begin{align}
%     & \bvarphi_{l-1} = \D_l \bvarphi_l, \nonumber \\
%     \text{s.t.} \;\; & \|\bvarphi_l\|_1\leq \lambda_l, \;\; \bvarphi_l \geq \mathbf{0}, \;\;\; \l = \{2,\dots,L\}\;,
%   \label{eq:msps}
% \end{align}
% \begin{align}
%     & \mathbf{s} = \D_1 \bvarphi_1,~~~\bvarphi_1 = \D_2 \bvarphi_2,\; \dots\;, ~~~\bvarphi_{L-1} = \D_L \bvarphi_L \;, \nonumber \\
%   & \forall l, ~~\|\bvarphi_l\|_1\leq \lambda_l, ~~ \bvarphi_l \geq \mathbf{0}.
%   \label{eq:msps}
% \end{align}
\begin{align}
%\resizebox{0.96\hsize}{!}{$
\resizebox{0.85\hsize}{!}{$
    \bvarphi_{l-1} = \D_l \bvarphi_l, \;\; \|\bvarphi_l\|_1\leq \lambda_l, \; \bvarphi_l \geq \mathbf{0}, \;\;\; \forall l\in\{1,\cdots,L\}\;,$}
\label{eq:msps}
\end{align}
where $\D_l \in \mathbb{R}^{K_{l-1}\times K_l}$ are the hierarchical dictionaries, $l$ is the index of hierarchy level, and $\lambda_l$ is the scalar specifying the amount of sparsity in each level. Thus the learnable parameters for $\C_\Theta$ is $\Theta=\{\cdots,\D_l,\lambda_l,\cdots\}$.
Constraints on multi-layer sparsity not only preserves sufficient freedom on shape variation, but it also results in more constrained code recovery.% and {\red a global dictionary}.

Multi-layer sparse coding induces a hierarchical block sparsity constraint on the block codes $\bPsi_l$ (equal to $\bvarphi_l \otimes \R_{xy}$ if orthogonal projection and $\bvarphi_l \otimes \R$ if perspective), which leads to a relaxation of the lower level problem in \eqref{eq:bilevel}:
% \begin{equation}
% \bPsi_{l-1} = (\D_l \otimes \I_3) \bPsi_l,~~~\|\bPsi_l\|_F^{(3\times a)}\leq \lambda_l, ~~\forall l,
% \label{eq:mblock}
% \end{equation}
% \begin{align}
%   &\M\tW = \M\tD_1 \tbPsi_1 \nonumber \\
%   \text{s.t.} \;\;\; & \|\bPsi_l\|_F^{(3 \times a)}\leq \lambda_l, \hspace{34pt} \l = \{1,\dots,L\} \nonumber \\
%   & \bPsi_{l-1} = (\D_l \otimes \I_3) \bPsi_l, \hspace{12pt} \l = \{2,\dots,L\} \;,
%   \label{eq:mbsparse_nrsfmpp}
% \end{align}
% \begin{equation}
%     \begin{aligned}
%  \min_{\{\bPsi_1,\cdots\bPsi_L\}} & \|\M\tW - \M\tD_1 \tbPsi_1\|_F \\
% {s.t.}~~~  & \|\bPsi_l\|_F^{(3 \times a)}\leq \lambda_l,\hspace{34pt}\forall l\in\{1,\cdots,L\},\\
%  & \bPsi_{l-1} = (\D_l \otimes \I_3) \bPsi_l,~~~~\forall l\in\{2,\cdots,L\},\\
%     \end{aligned}
%     \label{eq:mbsparse_nrsfmpp}
% \end{equation}
\begin{equation}
\resizebox{0.7\hsize}{!}{$
\begin{aligned}
 \min_{\{\bPsi_1,\cdots\bPsi_L\}} &  \|\M\tW - \M\tD_1 \tbPsi_1\|_F^2  + \sum_{l=1}^L \lambda_l\|\bPsi_l\|_F^{(3\times a)} \\ 
 & + \sum_{l=2}^L\|\bPsi_{l-1}-(\D_l\otimes\I_3)\bPsi_l\|_F^2 
\end{aligned}
$}
    \label{eq:mbsparse_nrsfmpp}
\end{equation}
where $\|.\|_F^{(3\times a)}$ denotes the sum of the Frobenius norm of each $3\times a$ block.
Here, $a=2$ for orthogonal projections and $a=3$ for perspective projections.
%{\red Given the definition of $\bPsi_1$, we can recover $\R$, $\bvarphi_1$ from the solution of~\eqref{eq:mbsparse_nrsfmpp}. ?}

\noindent\textbf{Remark.} Deep NR\SfM~\cite{ck19} further relaxes the block sparsity in~\eqref{eq:mbsparse_nrsfmpp} to $L_1$ sparsity with a nonnegative constraint to allow for the use of ReLU activations in the network architecture.
Such nonnegative constraint is inapplicable to our generalized formulation of $\bPsi$, and we empirically find the use of ReLU activation to degrade performance (Tab.~\ref{tab:cmu-mocap}).
% We find this relaxation unnecessary.
% In fact, the nonegative constraint does not apply to $\bPsi$ and would actually degrade empirical performance.

% =============================================================

\subsection{HBSC-induced Network Architecture}
\label{sec:deepnrsfm++}

The 2D-3D lifting network $f$ serves as an approximate solver of the HBSC problem~\eqref{eq:mbsparse_nrsfmpp}.
The architecture is induced from unrolling one iteration of the Iterative Shrinkage and Thresholding Algorithm (ISTA)~\cite{beck2009fast,rozell2008sparse,hastie2005elements}, one of the classic methods for sparse coding. Unrolling iterative solver as network architecture has been widely practised to insert inductive bias for better generalization~\cite{rick2017one,wang2018deep,tang2018ba,liu2017deep,Hu_2016_CVPR,chodosh2018deep,murdock2018deep}. However, the motivation here is different -- it is used to insert learnable priors to constrain an unsupervised learning problem.
% \cy{
% We choose to approximate ISTA which is one of the classic approach to solve the sparse coding problem. 
% }
% We start the derivation of the architecture by reviewing the block-sparse coding with a single layer.
We provide derivations in the following.

\vspace{6px}
\noindent \textbf{Block soft thresholding. }
% \subsection{ISTA with Block Soft Thresholding}
% \label{sec:bst}
We review the block-sparse coding problem and consider the single-layer case. To reconstruct an input signal $\mathbf{X}$, we solves:
\begin{equation}
    \min_{\bPsi} \|\mathbf{X} - \D \bPsi\|^2_F + \lambda\|\bPsi\|_F^{(3\times a)} \;. 
\end{equation}
One iteration of ISTA is computed as
\begin{equation}
    \everymath{\displaystyle}
    \resizebox{0.85\hsize}{!}{$
    \bPsi^{(t+1)} = \text{prox}_{\lambda\|\cdot\|_F^{(3\times a)}} ( \bPsi^{(t)}-\alpha \D^{\top}(\D\bPsi^{(t)}-\mathbf{X}) )
    $} \;,
\end{equation}
where $\text{prox}_{\lambda\|\cdot\|_F^{(3\times a)}}$ is the proximal operator for $L_1$ block sparsity of block size $3\times a$.
Let $\bPsi=[\Pi_1, \Pi_2, \dots \Pi_K]^{\top}$, where $\Pi_i \in \mathbb{R}^{3\times a} \; \forall i$.
Thus $\text{prox}_{\lambda\|\cdot\|_F^{(3\times a)}}$ is equivalent to applying \emph{block soft thresholding} (BST) to all $\Pi_i$, defined as
\begin{equation}
  \everymath{\displaystyle}
    \resizebox{0.85\hsize}{!}{$
    \text{BST}^{(3\times a)} (\bPsi; \lambda) = \begin{bmatrix} \dots & (1-\frac{\lambda}{\|\Pi_i\|_F})^+ \Pi_i^{\top}  & \dots  \end{bmatrix}^{\top}
    $} \;.
\end{equation}
Assuming the block code $\bPsi$ is initialized to $\mathbf{0}$ with the step size $\alpha=1$, the first iteration of ISTA can be written as
\begin{equation}
    \bPsi = \text{BST}^{(3\times a)} (\D^{\top}\mathbf{X}; \lambda) \;.
    \label{eq:block_ista}
\end{equation}
We interpret BST as solving for the block-sparse code and incorporate $\text{BST}^{(3\times a)}(\cdot)$ as the nonlinearity in our encoder part of the network, similar to a single-layer ReLU network interpreted as basis pursuit~\cite{papyan2017convolutional}.
% Similar to interpreting a single layer feed-forward network with ReLU as basis pursuit~\cite{papyan2017convolutional}, we can also interpret from~\eqref{eq:block_ista} that a network with block soft thresholding as nonlinearilty is solving for the block-sparse code. Based on this observation, we design the encoding part of our network to be a stack of layers with $\text{BST}^{(3\times M)}$ as nonlinearilty. 
Our formulation is closer to the true block-sparse coding objective than Deep NR\SfM, which uses ReLU as the nonlinearity to \emph{relax} the constraint to $L_1$ sparsity with nonnegative constraint.
% This is mainly due to the fact that the relationship $\bPsi = \bvarphi\otimes\mathbf{P}$ is not applicable to the non-negative constraint.
% We show empirical evidence of the superiority of $\text{BST}^{(3\times a)}$ over ReLU in Table~\ref{tab:cmu-mocap}.
% In Deep NR\SfM, instead of enforcing $L_1$ block sparsity, it relaxes the constraint to $L_1$ sparsity with nonegative constraint. This leads to replacing $\text{BST}^{(3\times M)}$ with ReLU. However we argue that this relaxation in theory is undesirable, because not only it departs from the true objective of block-sparse coding, but also $\bPsi = \bvarphi\otimes\mathbf{P}$ is not applicable to the nonegative constraint. Empirical results (see Table~\ref{tab:cmu-mocap}) also support the use of $\text{BST}^{(3\times M)}$ over ReLU.

\vspace{6px}
\noindent \textbf{Encoder-decoder network. }
By unrolling one iteration of block ISTA for each layer, our encoder takes $\tW$ as input and produces the block code for the last layer $\bPsi_L$ as output:
\begin{equation}
\begin{aligned}
\bPsi_1 = &\text{BST} ^{(3\times a)} \left(
[\tD_1^{\top} \tW]_{3K_1 \times a}; \blambda_1  \right),\\
\bPsi_2 = & \text{BST} ^{(3\times a)} \left(
(\D_2 \otimes \I_3)^{\top} \bPsi_1;  \blambda_2 \right),\\
&\vdots\\
\bPsi_L = & \text{BST} ^{(3\times a)} \left(
(\D_L \otimes \I_3)^{\top} \bPsi_{L-1};  \blambda_L \right),
\end{aligned}
\end{equation}
where $\blambda_l$ is the learnable threshold for each $K_l$ block and $[\cdot]_{3K_1\times a}$ is a $3K_1\times a$ reshape. $(\D_l\otimes\I_3)^T\bPsi_{l-1}$ are implemented by convolution transpose.
% We have tried analytical solutions as well as approximate solutions with a single layer networks. Both give similar results. Therefore 
$\bPsi_L$ are then factorized into $\bvarphi_L$, $\R$ (constraining to $\SO(3)$ using SVD~\cite{ck19}).
The 3D shape $\mathbf{S}$ is recovered from $\bvarphi_L$ via the decoder as:
% The code $\bvarphi_L$ is passed into the decoder to recover the 3D shape $\mathbf{S}$ via
\begin{equation}
\begin{aligned}
\bvarphi_{L-1} &= \text{ReLU}(\D_L \bvarphi_{L} + \mathbf{b}_L) \;,  \\
\;\; & \vdots \\
\bvarphi_{1} &= \text{ReLU}(\D_2 \bvarphi_{2}  + \mathbf{b}_2) \;,\\
\S &= \D_1 \bvarphi_{1} \;. \\
\end{aligned}
\end{equation}

\noindent\textbf{Remark.}
Our key technical differences to Deep NR\SfM are: (i) the first layer of the network is adaptive according to the camera model and keypoint visibility; (ii) replacement of ReLU with BST; (iii) block size of $\bPsi_l$ becomes $3\times3$ under the perspective camera model.

% \paragraph{Assumptions}
% We make the following the assumptions: i) camera intrinsics are known. With this, we simplify the problem by assuming the intrinsic matrix $\mathbf{K} = \I_3$; (ii) the distance between the object center and camera is given; (iii) each input 2D keypoints are observed, occlusions would be handled in the next section. (iv) 3D shape can be decomposed as a linear combination of shape basis, and the coefficients are assumed to be constrained by a multi-level sparse coding objective.

% \paragraph{Notations} Define basic notation here.

%% file: sections/5-experiment.tex
\section{Experiments}
\label{sec:exp}
% \vspace{6px}

% \begin{figure}[t!]
%     \centering
%     \includegraphics[width=1\linewidth]{figures/datasets.pdf}
%     \caption{Qualitative results across different datasets. Blue points: ground truth, yellow points and orange lines: reconstruction. Red lines: difference between ground truth and reconstruction.}
%     \label{fig:datasets}
% \end{figure}

\noindent \textbf{Architectural details. }
% The only hyperparameters for our approach are the number of layers and the size of the dictionaries.
% Our model is much more compact compared to residual networks~\cite{he2016deep} used by other approaches.
The most important hyperparameter of Deep NR\SfMpp is the dictionary size $K_L$ at the last level, which depends on the shape variation that exhibits within the dataset. The rest are set arbitrarily and has less affect on performance. Setting $K_L$ to 8 gives reasonable result across all evaluated tasks. For optimal performance by validating on hold-out validation set, we use 8-10 for articulated objects and 2-4 for rigid ones. The detailed description of the architecture and the analysis of the robustness of $K_L$ are included in the Supp.
% For optimal performance, we set it to 8-10 for articulated objects and 2-4 for rigid objects.
% Our model is much more compact compared to residual networks~\cite{he2016deep} used by other approaches.
% the residual networks used in other deep learning approaches, our model is much more compact.
% We use Adam~\cite{kingma2014adam} to train our method. Detailed description is provided in supplementary.

\vspace{6px}
\noindent \textbf{Shape scale $t_z$. }
To set the scale $t_z$ in practice, we make use of available 2D information such as bounding boxes; in applications where relative scales between samples in a dataset is available (\eg human skeletons), we can utilize a strategy to estimate and recorrect the scale $t_z$ in an iterative fashion.
% The most proper way to normalize the scale of 2D inputs is by using the ground truth $t_z$ if such oracle information is available. Although in some applications, $t_z$ with metric scale can be obtained by depth sensing or 3D localization, in this work we keep our algrithm as general as possible, therefore we do not assume any oracle 3D information is accessible.
We use the detected 2D object bounding box to provide an initial estimate of $t_z$ and subsequently update the scale estimation (using the Frobenius norm of $\S$ or the average bone length of a skeleton model, if available).
Once we have updated the scale estimation $t_z$, we rerun Deep NR\SfMpp~and update the reconstruction.
This \textbf{scale correction} procedure allows the 3D reconstruction and scale estimation to improve each other.

\vspace{6px}
\noindent \textbf{Evaluation metrics. }
We employ the following metrics to evaluate the accuracy of 3D reconstruction. \textbf{MPJPE}: before calculating the mean per-joint position, we normalize the scale of the prediction to match against ground truth (GT). To account for the ambiguity from weak perspective cameras, we flip the depth value of the prediction if it leads to lower error. \textbf{PA-MPJPE}: rigid align the prediction to GT before evaluating MPJPE. \textbf{STRESS}: borrowed from Novotny \etal~\cite{c3dpo} is a metric invariant to camera pose and scale. \textbf{Normalized 3D error}: 3D error normalized by the scale of GT, used in prior NR\SfM works~\cite{akhter2009nonrigid,dai2014simple,ck19,gotardo2011kernel}.

\begin{table}[t!]
    \centering
    % \footnotesize
    \setlength\tabcolsep{4pt}
    % \begin{small}
    \resizebox{0.92\linewidth}{!}{
    \begin{tabular}{c|cccc|cc}
    \hline
      \small{Subject} & \small{CNS} & \small{NLO} & \small{SPS} & \small{Deep NR\SfM} & \small{ReLU}  &  \small{BST}\\
    %   Subject  & CNS~\cite{lee2016consensus} & NLO~\cite{del2007non} & SPS~\cite{kong2016prior} & Deep NR\SfM~\cite{ck19} & ReLU~\cite{ck19} &  Deep NR\SfMpp\\
       \hline
        1 & 0.613 & 1.22 & 1.28 & 0.175 & 0.265 & \textbf{0.112}\\ 
        5 & 0.657 & 1.160 & 1.122 & \textbf{0.220} & 0.393 & 0.230\\
        18 & 0.541 & 0.917 & 0.953 & 0.081 & 0.117 & \textbf{0.076}\\
        23 & 0.603 & 0.998 & 0.880 & 0.053 & 0.093 & \textbf{0.048}\\
        64 & 0.543 & 1.218 & 1.119 & 0.082 & 0.179 & \textbf{0.020}\\
        70 & 0.472 & 0.836 & 1.009 & 0.039 & 0.030 & \textbf{0.019}\\
        % 102 & 0.581 & 1.144 & 1.087 & 0.115 & 0.581 & 0.485\\
        106 & 0.636 & 1.016 & 0.957 & \textbf{0.113} & 0.364 & 0.116\\
        123 & 0.479 & 1.009 & 0.828 & 0.040 & 0.040 & \textbf{0.020}\\
        % 127 & 0.644 & 1.050 & 1.025 & 0.095 & 0.097 & 0.572\\
    \hline
    \end{tabular}
    % \vspace{4pt}
    }
    \caption{3D reconstruction error on CMU Motion Capture compared with NR\SfM methods: CNS~\cite{lee2016consensus}, NLO~\cite{del2007non}, SPS~\cite{kong2016prior} and Deep NR\SfM~\cite{ck19}. ``ReLU'' and ``BST'' are Deep NR\SfM reimplemented with different nonlinearities.}
    \label{tab:cmu-mocap}
    % \vspace{-8pt}
\end{table}

\vspace{6px}
\noindent \textbf{BST vs ReLU. }
To study the effect of replacing ReLU with BST as the nonlinearity, we compare our approach against Deep NR\SfM~\cite{ck19} on orthogonal projection data with perfect point correspondences on the CMU motion capture dataset~\cite{cmumocap}.
Normalized 3D error is reported per human subject in Table~\ref{tab:cmu-mocap}.
Our approach using BST achieves better accuracy compared to Deep NR\SfM, providing empirical benefits of its closer proximity to solving the true block sparse coding objective.
% with closer proximity to solving the true block sparse coding objective, namely using block soft thresholding instead of ReLU, achieves better accuracy compared to our own re-implementation (ReLU) of Deep NR\SfM as well as the numbers reported in the original paper.
%It also shows dramatic improvements comparing against other NR\SfM methods.
% Table~\ref{tab:cmu-mocap} shows that our approach with closer proximity to solving the true block sparse coding objective, namely using block soft thresholding instead of ReLU, achieves better accuracy compared to our own re-implementation (ReLU) of Deep NR\SfM as well as the numbers reported in the original paper. We also compare against other NR\SfM methods and show dramatic improvement. 

\vspace{6px}
\noindent \textbf{Orthogonal projection with missing data. }
We evaluate Deep NR\SfMpp on two benchmarks with high amount of missing data (Table.~\ref{tab:pascal}):
(1) Synthetic UP-3D, a large synthetic dataset with dense human keypoints based on the UP-3D dataset~\cite{up3d}.
The data was generated by orthographic projections of the SMPL body shape with the visibility computed from a ray tracer.
% The goal is to reconstruct 3D shapes from the rendered 2D keypoints.
We follow the same settings as C3DPO~\cite{c3dpo} and evaluate the 3D reconstruction of 79 representative vertices of the SMPL model on the test set.
(2) PASCAL3D+~\cite{xiang2014beyond} consists of images of 12 rigid object categories with sparse keypoints annotations. To ensure consistency between 2D keypoint and 3D ground truth, we follow C3DPO and use the orthographic projections of the aligned CAD models with the visibility taken from the original 2D annotations. A single model is trained to account for all 12 object categories.
We also include results where the 2D keypoints are detected by HRNet~\cite{hrnet}.

\begin{table*}[t!]
    \centering
    \begin{subtable}{0.3\linewidth}
        \centering
        \setlength\tabcolsep{4pt}
        \resizebox{\linewidth}{!}{
    \begin{tabular}{c|c|c}
    \hline
        Method & MPJPE & STRESS \\
    \hline
        EM-SfM~\cite{torresani2008nonrigid} & 0.107 & 0.061\\
        GbNRSfM~\cite{fragkiadaki2014grouping} & 0.093 & 0.062\\
        Deep NRSfM~\cite{ck19} & 0.076 & 0.063 \\
        C3DPO-base~\cite{c3dpo} & 0.160 & 0.105\\
        C3DPO~\cite{c3dpo} & 0.067 & 0.040\\
        Deep NR\SfMpp & \textbf{0.062} & \textbf{0.037} \\
    \hline
    \end{tabular}    
        }
        \caption {Test error on Synthetic UP-3D.}
    \end{subtable}
    \hspace{8pt}
    \begin{subtable}{0.3\linewidth}
        \centering
        \setlength\tabcolsep{4pt}
        \resizebox{\linewidth}{!}{
        \begin{tabular}{c|c|c}
            \hline
            Method & MPJPE & STRESS \\
            \hline
            EM-SfM~\cite{torresani2008nonrigid} & 131.0 & 116.8 \\
            GbNRSfM~\cite{fragkiadaki2014grouping} & 184.6 & 111.3\\
            Deep NR\SfM~\cite{ck19} & 51.3 & 44.5 \\
            C3DPO-base~\cite{c3dpo} & 53.5 & 46.8\\
            C3DPO~\cite{c3dpo} & 36.6 & 31.1\\
            Deep NR\SfMpp & \textbf{34.8} & \textbf{27.9} \\
            \hline
        \end{tabular}
        }
        \caption{Test error on PASCAL3D+, where the input keypoints are from the ground truth.}
    \end{subtable}
    \hspace{8pt}
    \begin{subtable}{0.3\linewidth}
        \centering
        \setlength\tabcolsep{4pt}
        \resizebox{1\linewidth}{!}{
        \begin{tabular}{c|c|c}
            \hline
            Method & MPJPE & STRESS \\
            \hline
            Deep NR\SfM~\cite{ck19} & 65.3 & 47.7 \\
            CMR~\cite{cmr} & 74.4 & 53.7 \\
            C3DPO~\cite{c3dpo} & 57.5 & 41.4 \\
            Deep NR\SfMpp & \textbf{53.0} & \textbf{36.1} \\
            \hline
        \end{tabular}
        }
        % \label{tab:pascal}
        \caption{Test error on PASCAL3D+, where the input keypoints are off-the-shelf keypoint detection results from HRNet~\cite{hrnet}.}
    \end{subtable}
    \vspace{-4pt}
    \caption{Quantitative results on data generated by orthogonal projection with realistic occlusions.}
    \label{tab:pascal}
    % \vspace{-8pt}
\end{table*}

Our method achieves over 32\% error reduction over Deep NR\SfM (Table~\ref{tab:pascal}(b)) while comparing favorably against other NR\SfM methods and deep learning methods such as C3DPO.

\begin{table*}[t!]
    \centering
    \begin{subtable}{0.58\linewidth}
    \centering
    \small
    \setlength\tabcolsep{4pt}
    \resizebox{\linewidth}{!}{
        \begin{tabular}{c|cccc||c|ccc}
        \hline
        \multirow{2}{*}{Method} &
        \multirow{1}{*}{KSTA} &
        \multirow{1}{*}{CNS} & 
        \multirow{1}{*}{Pose-GAN} & 
        \multirow{1}{*}{C3DPO} & 
        \multirow{1}{*}{Chen~\etal} & 
        \multicolumn{3}{c}{Deep NR\SfMpp} \\
         & \cite{gotardo2011kernel} & \cite{lee2016consensus} & \cite{posegan} & \cite{c3dpo} 
         & \cite{chen_drover}
         & \multirow{2}{*}{ortho.} & \multirow{2}{*}{persp.} & \scriptsize persp. + \vspace{-4pt} \\
         & & & & & & & & \scriptsize scale corr. \\
        \hline
        MPJPE & - & 120.1 & 130.9 & 95.6 & - & 104.2 & 60.5 & \bf 56.6 \\
        PA-MPJPE & 123.6 & 79.6 & - & - & 58 & 72.9 & 51.8 & \bf 50.9 \\
        \hline
        \end{tabular}
    }
    \vspace{-4pt}
    \caption{Test error on Human 3.6M compared against unsupervised methods. We report our results under both weak perspective and perspective camera models and demonstrate the effectiveness of applying scale corrections (2 iterations).}
    % The different versions of our method are: + persp: using perspective projection model, +scale corr itr: applying different number of scale correction iterations.}
    \label{tab:h36m}
    \end{subtable}
    \hspace{8pt}
    \begin{subtable}{0.38\linewidth}
        \centering
        \setlength\tabcolsep{4pt}
        \resizebox{\linewidth}{!}{
            \begin{tabular}{cc|c|cc|cc}
            \hline
            \multicolumn{3}{c}{} & \multicolumn{2}{c}{w/o missing pts.} &  \multicolumn{2}{c}{w/ missing pts.}\\
            \multicolumn{3}{c}{method} & train & test  & train & test \\
            \hline
            \multicolumn{3}{c}{CNS~\cite{lee2016consensus}} & 1.30 & - & - & - \\
            \multicolumn{3}{c}{KSTA~\cite{gotardo2011kernel}} & 1.58 & - & 1.62 & - \\
            \hline
            \multirow{3}{*}{\rotatebox{90}{Deep}} \hspace{-8pt} & \multirow{3}{*}{\rotatebox{90}{\footnotesize NR\SfMpp}} & ortho. & 0.596 & 0.591 & 0.679 & 0.681\\
            && persp. & 0.152 & 0.145 & 0.182 & 0.185\\
            && + scale correction & 0.131 & 0.124 & 0.165 & 0.168\\
            \hline
            \end{tabular}
        }
        \vspace{-4pt}
        \caption{Test error on ApolloCar3D, where we report the MPJPE in meters as the evaluation metric.}
        \label{tab:apollo1}
    \end{subtable}
    \vspace{-4pt}
    \caption{Quantitative results on real-world data with perspective projection.}
\end{table*}

% \begin{table}[t!]
%     \centering
%         \centering
%         \small
%             \begin{tabular}{c|c|c|c}
%             \hline
%                 & \multicolumn{3}{c}{Missing pts. (\%)}\\
%             Noise($\sigma$)    & 0 & 30 & 60 \\
%             \hline
%             0    & 0.124 & 0.142 & 0.192\\
%             3    & 0.129 & 0.144 & 0.205\\
%             5    & 0.136 & 0.150 &  0.202\\
%             10   & 0.125 & 0.166 & 0.181\\
%             15   & 0.191 & 0.188 & 0.304\\
%             \hline
%             \end{tabular}
%         \vspace{-4pt}
%         \caption{Robustness with input under different occlusion rates and noises on \textbf{ApolloCar3D} dataset.}
%         \label{tab:noise}
%     % \vspace{-8pt}
% \end{table}

\vspace{6px}
\noindent \textbf{Perspective projection.}
We evaluate our approach on two datasets with strong perspective effects:
(1) Human 3.6M~\cite{h36m}, a large-scale human pose dataset annotated by motion capture systems. We closely follow the commonly used evaluation protocol: we use 5 subjects  (1, 5, 6, 7, 8) for training and 2 subjects (9, 11) for testing.
(2) ApolloCar3D~\cite{song2019apollocar3d} has 5277 images featuring cars, where each car instance comes with annotated 3D pose.
% was annotated with 3D pose by running PnP with 3D CAD models.
2D keypoint annotations are also provided without 3D ground truth. To evaluate our method, we render 2D keypoints by projecting 34 car models according to the 3D pose labels.
Visibility of each keypoints are marked according to the original 2D keypoint annotations.
To showcase strong perspective effects, we select cars within 30 meters with no less than 25 visible points (out of 58 in total), which gives us 2848 samples for training and 1842 for testing.

\begin{figure*}[t!]
    \centering
    \includegraphics[width=1\linewidth]{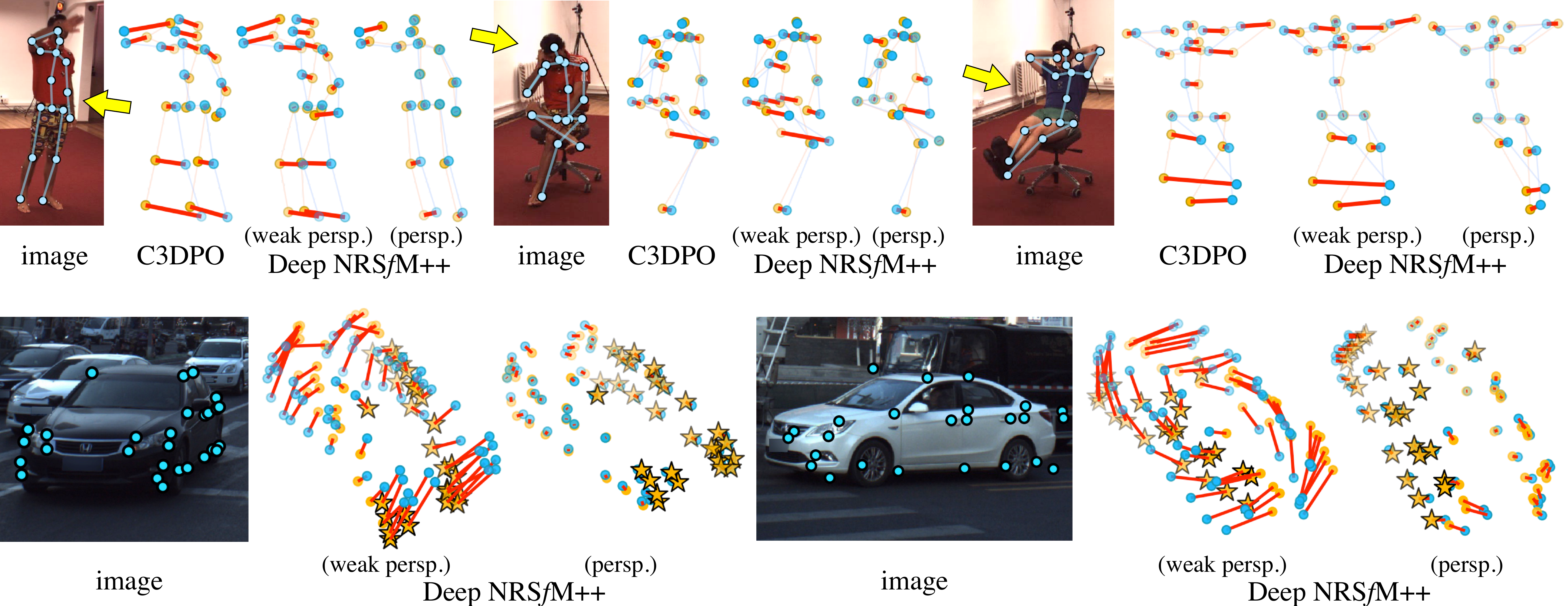}
    \caption{Qualitatative comparison.
    Blue points: GT, yellow points: prediction (stars indicate visible annotations), red lines: prediction error.
    The two rows are visual results from Human 3.6M and ApolloCar3D datasets respectively.
    }
    \label{fig:qualitative}
\end{figure*}

We evaluate different variants of our approach. We find that modeling perspective projection (Deep NR\SfMpp persp.) leads to significant improvement over the orthogonal model (Deep NR\SfMpp ortho.) and applying scale correction further improves accuracy.
Deep NR\SfMpp shows robustness under different level of noise and occlusion (see Table~\ref{tab:noise}) and achieves the best result compared to other unsupervised learning method.
We outperform the leading GAN-based method~\cite{chen_drover} by a significant margin when using the same training set (50.9 v.s. 58); in addition, Chen~\etal~\cite{chen_drover} reaches our level of performance (50.9 vs 51) only when external training sources and temporal constraints are used.
We provide qualitative results in Fig.~\ref{fig:qualitative}, which shows the benefit of Deep NR\SfMpp's ability to model perspective camera models while also outperforming C3DPO.

\begin{table*}[t!]
    \centering
    \parbox{0.25\linewidth}{
        \resizebox{\linewidth}{!}{
        \begin{tabular}{c|c|c|c}
            \hline
            Noise & \multicolumn{3}{c}{Missing pts. (\%)}\\
            ($\sigma$) & 0 & 30 & 60 \\
            \hline
            0    & 0.124 & 0.142 & 0.192\\
            3    & 0.129 & 0.144 & 0.205\\
            5    & 0.136 & 0.150 &  0.202\\
            10   & 0.125 & 0.166 & 0.181\\
            15   & 0.191 & 0.188 & 0.304\\
            \hline
        \end{tabular}
        \vspace{-4pt}
        }
        \caption{Robustness with input under different occlusion rates and noises on ApolloCar3D.}
        \label{tab:noise}
    }
    \hspace{8px}
    \parbox{0.72\linewidth}{
        \centering
        \includegraphics[width=1\linewidth]{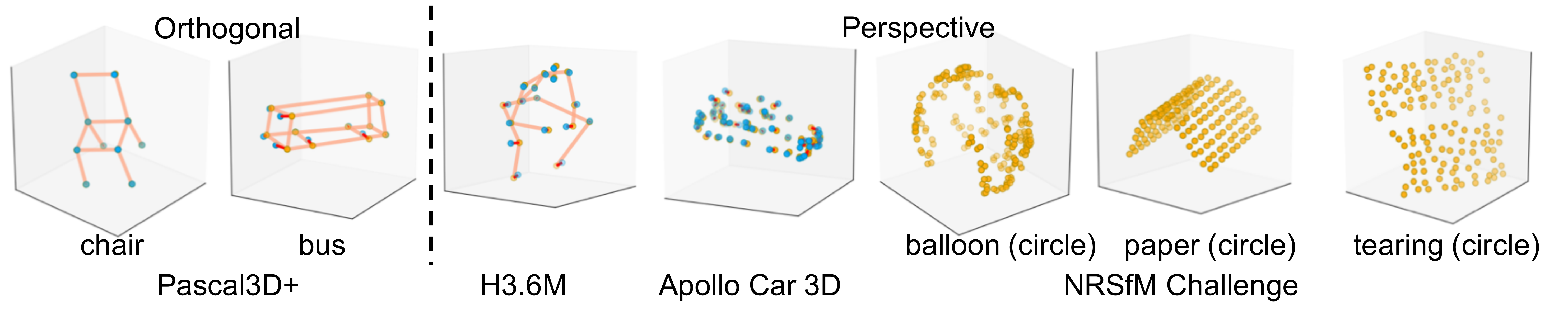}
        \caption{Qualitative results across different datasets. Blue points: ground truth; yellow points and orange lines: reconstruction; red lines: difference between ground truth and reconstruction. Best viewed in color and zoomed in.}
        \label{fig:datasets}
    }
    \vspace{-4pt}
\end{table*}

\vspace{6px}
\noindent \textbf{Deforming object in short sequences.} Our method is applicable to NR\SfM problems with limited number of frames. We experiment on \textbf{NR\SfM Challenge Dataset}~\cite{jensen2018benchmark}, which consists of 5 deforming objects captured with six different camera paths. While the leading methods all utilize trajectory information, our atemporal approach still gives reasonable reconstruction as shown in Fig.~\ref{fig:datasets}. The detailed comparison is in the supp. material.
%(Articulated, Balloon, Paper, Stretch \& Tearing)(\ie circle, flyby, line, semi-circle, tricky, zig-zag

%% file: sections/6-conclusion.tex
\section{Conclusion}
% We propose Deep NR\SfMpp, an atemporal approach applicable to SfC and unsupervised pose estimation. It models both weak and strong perspective camera as well as missing data in an elegant biliniar factorization form. We demonstrate state-of-the-art performances across numerous benchmarks against various NR\SfM techniques as well as deep learning based approaches, indicating the effectiveness of our approach. The \textbf{limitations} we find in this work is: (i) the hierarchical sparsity prior is less robust to articulated objects when only limited number of frames is available (see Supp.). (ii) although we establish a closer theoretical connection of an end-to-end learning framework to the true objective of the NR\SfM problem, this connection is still an approximation. Further theoritical investigation is in need to understand the benifit of this approximation or any possible alternatives, but is out of the scope of this paper.

 % Furthermore, we establish a closer theoretical connection of an end-to-end learning framework to the true objective of the NR\SfM problem.

 We propose an atemporal approach applicable to SfC and unsupervised pose estimation. It provides an unified framework to model both orthogonal and perspective camera as well as missing data. The \textbf{limitations} of this work is: (i) the hierarchical sparsity prior is less robust to articulated objects when only limited number of frames is available (see Supp.). (ii) the connection between the proposed learning framework and the true objective of the NR\SfM problem is still an approximation. Further theoritical investigation is in need to understand the benifit of this approximation or any possible alternatives, but is out of the scope of this paper.

%% file: sections/supp.tex
% \section{Typo correction}
% In Equation (\textcolor{red}{6}), $\|\bPsi_i\|_1^{(3\times M)}$ should be $\|\bPsi_i\|_F^{(3\times M)}$.
% \section*{Appendix}

\subsection*{I: Derivation for handling missing data under orthogonal camera. }
To take possible occlusions into account, we have
\begin{equation}
    \M \W = \M (\S \R_{xy} + \mathbf{1}_P  \mathbf{t}_{xy}^{\top}) \;.
    \label{eq:orth_eq_occ}
\end{equation}
where $\M = \text{diag}(m_1, \cdots, m_P)$ is a diagonal matrix indicating the visibility for each keypoints, $\R_{xy}$ is the first two columns in rotation matrix $\R$. By the object centric constraint, $\mathbf{t}_{xy}$ has the followin equation:
\begin{equation} \label{eq:newt}
    \mathbf{t}_{xy} = \frac{1}{P}\sum_{i=1}^P \underbrace{m_i \mathbf{w}_i}_{\text{visible points}} + \underbrace{(1-m_i)(\R_{xy}^{\top}\mathbf{s}_i + \mathbf{t}_{xy})}_{\text{occluded points}},
\end{equation}
where $m_i$ indicates the visibility of the $i$-th keypoint and $\mathbf{s}_i$ denotes the $i$-th 3D point in $\S$.
Rearranging~\eqref{eq:newt} yields an expression of $\t_{xy}$ with unknowns only from $\R_{xy}$, $\S$:
% By moving all the terms with $\t_{xy}$ to one side of the equation, we can express $\t_{xy}$ as:
\begin{align} \label{eq:rearrange}
    \mathbf{t}_{xy} &=  \frac{1}{\Tilde{P}}\sum_{i=1}^P m_i \mathbf{w}_i + (1-m_i) \R_{xy}^{\top}\mathbf{s}_i \nonumber \\
    &= \frac{1}{\Tilde{P}} [ \M \W + (\I_P - \M)\S\R_{xy}]^{\top}\mathbf{1}_P \;,
\end{align}
where $\tilde{P}$ denotes the number of visible points.
Substituting~\eqref{eq:rearrange} into~\eqref{eq:orth_eq_occ} and rearranging, we have
\begin{equation}
\footnotesize{
    \M (\W - \mathbf{1}_P  \frac{\mathbf{1}_P^{\top} \M\W}{\Tilde{P}}  ) = 
    \M(\S\R_{xy} + \mathbf{1}_P  \frac{\mathbf{1}_P^{\top} (\I_P - \M)\S\R_{xy}}{\Tilde{P}}) \;.
}
\end{equation}
Since $\S\R_{xy} = \D^\sharp (\bvarphi \otimes \R_{xy}) =  \D^\sharp \bPsi$, we have
\begin{equation}
\footnotesize{
    \M  \underbrace{ (\W - \mathbf{1}_P \frac{\mathbf{1}_P^{\top} \M\W}{\Tilde{P}}}_{\Tilde{W}\text{: normalized 2D projection}}  ) = 
    \M \underbrace{(\D^\sharp + \mathbf{1}_P  \frac{\mathbf{1}_P^{\top} (\I_P - \M)\D_1^\sharp}{\Tilde{P}})}_{\Tilde{\D} \text{: normalized dictionary}}\bPsi
    }
\label{eq:orth_occlu}
\end{equation}
% In other words, $\tilde{\W}$ is formed by shifting $\W$ with the average of visible keypoints locations.
% This aligns with the common practice employed for data normalization~\cite{ck19, c3dpo}. 
% For the special case where all points are visible, the expressions in~\eqref{eq:orth_occlu} degenerates back to~\eqref{eq:mblock}. 

\subsection*{II: Implementation details}
\begin{figure}
    \centering
    \includegraphics[width=.7\linewidth]{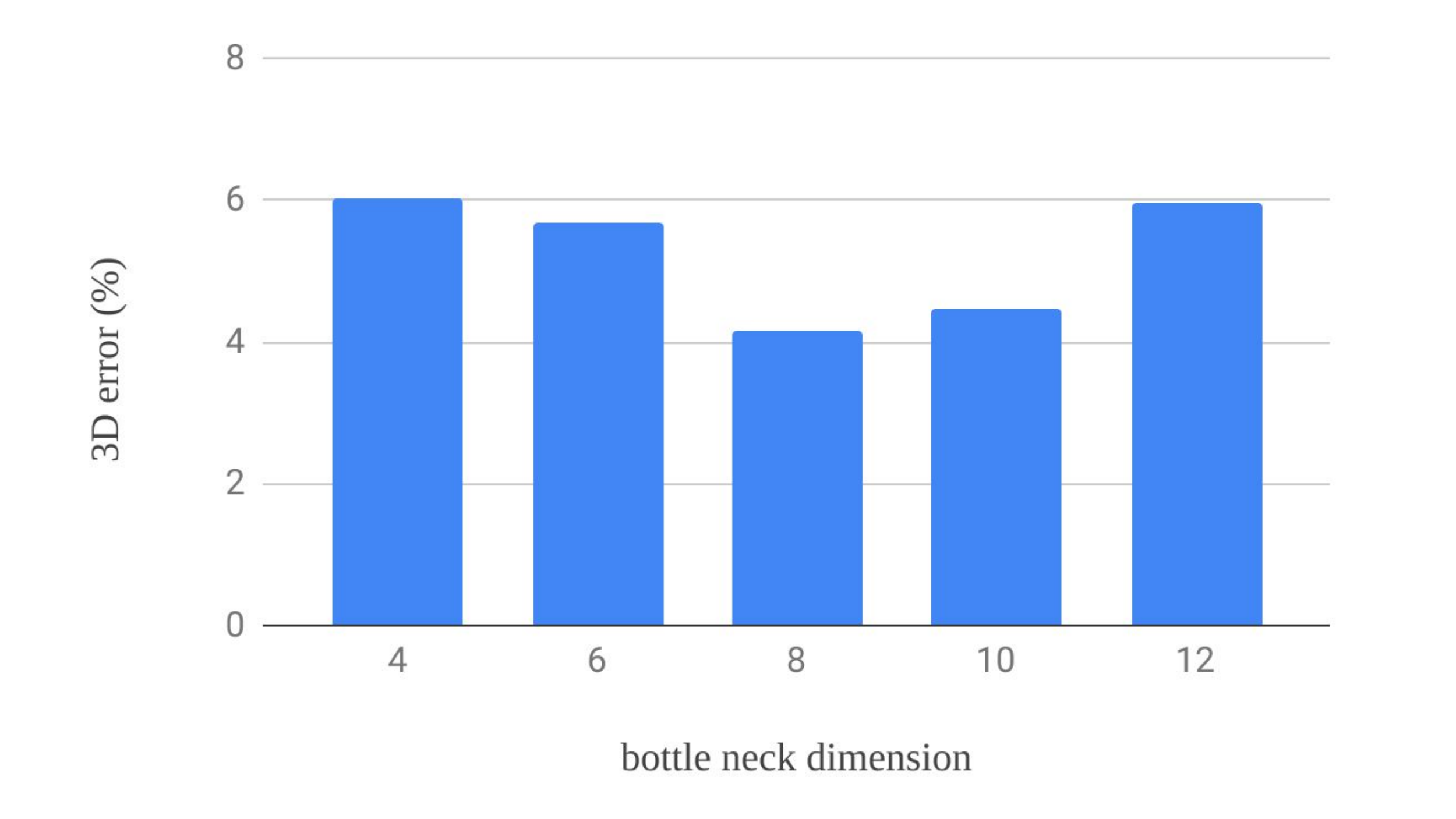}
    \caption{3D reconstruction accuracy with different bottleneck dimensions on CMU-Mocap subject 64.}
    \label{fig:my_label}
\end{figure}

\paragraph{Dictionary sizes.}
The dictionary size in each layer of the block sparse coding is listed in Table~\ref{tab:dict_size}. We tried two strategies to set the dictionary sizes: (i) exponentially decrease the dictionary size, i.e. 512, 256, 128, ..., (ii) linearly decrease, i.e. 125, 115, 104, ... . Both strategies give reasonably good results. However, the hack we need to perform is to pick the size of the first and last layer dictionaries. We find that the size of the first layer would not have a major impact on accuracy as long as it is sufficiently large. The major performance factor is the size of the last layer dictionary $K_L$. In principal, we shall pick $K_l$ by measuring the accuracy on a small holdout validation set with 3D ground truth. We admit that this may become a problem in practise when no validation information is available. Indeed, having a more generalizable solution compared to hyperparameter tuning would be of interest to not only this work but also many other NRSfM or unsupervised 3D pose lifting approaches, which is an important topic to explore in the future. For now, we rely on the robustness of our approach against different bottleneck dimensions, and the rest hyperparameters are set arbitrarily without tuning. Figure~\ref{fig:my_label} shows the 3D accuracy with different bottleneck dimensions on CMU-Mocap subject 64. It shows optimal result at dimension around 8 and 10, and has reasonable result at both lower and higher dimensions.

%and the corresponding hierarchical dictionary is deep enough (e.g. $256+$ for exponential decrease, $125+$ for linear decrease). The major performance factor is the size of the last layer dictionary. Throughout the datasets we experimented, we found setting this scalar as 8 gets reasonable performance. To get optimal result, the rule of thumb we discovered is: 2-4 for rigid objects and 8-10 for articulated objects such as human body.

%In principal, we shall pick the hyperparameters by measuring the accuracy on a small holdout validation set with 3D ground truth. We admit that this may become a problem in practise when no validation information is available. Indeed, having a more generalizable solution compared to hyperparameter tuning would be of interest to not only this work but also many other NRSfM or unsupervised 3D pose lifting approaches, which is an important topic to explore in the future.\\
%For now, we rely on the robustness of our approach against different bottleneck dimensions, and the rest hyperparameters are set arbitrarily without tuning. Figure~\ref{fig:my_label} shows the 3D accuracy with different bottleneck dimensions on CMU-Mocap subject 64. It shows optimal result at dimension around 8 and 10, and has reasonable result at both lower and higher dimensions.  More results with different object classes would be included.  

\paragraph{Training parameters.}
We use Adam optimizer to train. Learning rate = 0.0001 with linear decay rate 0.95 per 50k iterations. The total number of iteration is 400k to 1.2 million depending on the data. Batch size is set to 128. Larger or smaller batch sizes all lead to similar result. 

\paragraph{Initial scale for input normalization.}
For weak perspective and strong perspective data, we need to estimate the scale so as to properly normalize the size of the 2D input shape. For Pascal3D+ and H3.6M datasets, we use the maximum length of the 2D bounding box edges, i.e. $t_z = 1/\max( \text{bbox\_height, bbox\_width})$. For Apollo 3D Car dataset, we choose the minimum length of the bounding box, i.e. $t_z = 1/\min( \text{bbox\_height, bbox\_width})$ by assuming that the height of each car is identical.

\subsection*{III: Additional empirical analysis}
% \paragraph{Pascal3D+}
\paragraph{H3.6M.} We add the test result using detected 2D keypoint as input in Table~\ref{tab:h36m_cpn}. Deep NR\SfM++ achieves state-of-the-art result compared to other unsupervised methods.

\begin{figure}[h]
    \centering
    \begin{tabular}{*{2}{c@{\hspace{3px}}}}
    \includegraphics[width=.48\linewidth]{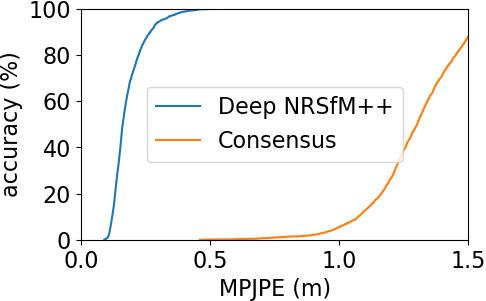} &
    \includegraphics[width=.48\linewidth]{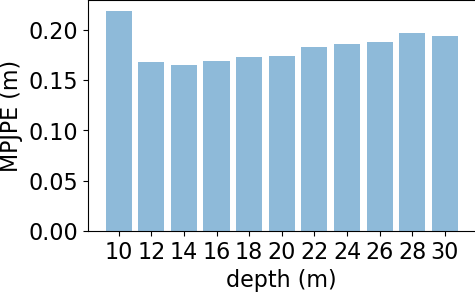} \\
    \includegraphics[width=.48\linewidth]{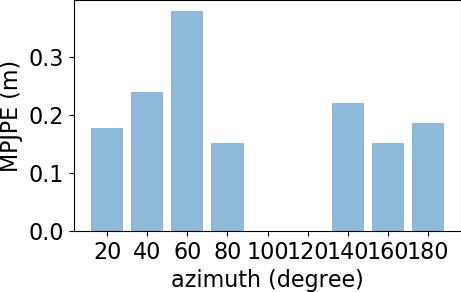} &
    \includegraphics[width=.48\linewidth]{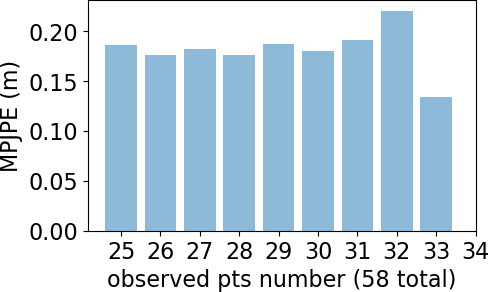}    
    \end{tabular}
    \caption{Additional result on \textbf{Apollo 3D Car} dataset. Top-left: percentage of sucess at different error thresholds. Rest: average error at different distances to the camera (top-right), azimuth angles of the car (bottom-left) and number of observed keypoints (bottom-right).}
    \label{fig:apollo_supp}
\end{figure}

\paragraph{ApolloCar3D.} Figure~\ref{fig:apollo_supp} shows additional analysis of Deep NR\SfM++ on the training set. Our method achieves $< 25$cm error for over 80\% testing samples with occlusions, while the compared baseline method, namely Consensus NR\SfM~\cite{lee2016consensus} fails to produce meaningful reconstruction using perfect point correspondences. Average errors at different distances, rotation angles and occlusion rates are also reported. Overall, our method does not have a strong bias against a particular distance or occlusion rate. It does show larger error at $60^\circ$ azimuth, most likely due to the data distribution, where most cars are in either front ($\approx0^\circ$) or back ($\approx180^\circ$) view.

\paragraph{NRSfM Challenge Dataset.} The leading methods on the leaderboard all utilize temporal constraints, while our \textbf{atemporal} approach solves the problem only in the shape space. Therefore it is natural for our method to fall behind those methods, but nonetheless, we achieve reasonable results compared to other approaches also only use shape constraints. In Table~\ref{tab:chall_orth} and Table~\ref{tab:chall_persp}, we compare to a selection of classical methods in literature. For a complete comparison, please refer to the official leaderboard of the challenge~\cite{jensen2018benchmark}.

\subsection*{IV: Additional discussion}
One of the benefit of solving NR\SfM by training a neural network is that, in addition to 2D reconstruction loss, we can easily employ other loss functions to further constrain the problem. In summary, our preliminary study finds that: (i) adding the canonicalization loss~\cite{c3dpo} does not noticeably improve result. (ii) adding Lasso regularization on $\bvarphi_1$ gives marginally better result in some datasets. (iii) adding symmetry constraint on the skeleton bone length helps to improve robustness against network initialization, but does not lead to noticeable better accuracy. 

% \begin{table}[]
%     \centering
%     \begin{tabular}{c|cc|cc}
%     \hline
%          & \multicolumn{2}{c}{C3DPO~\cite{c3dpo}} & \multicolumn{2}{c}{Deep NR\SfMpp}\\
%         & MPJPE & Stress & MPJPE & Streess\\
%     \hline
% aeroplane & 17.4 & 12.7 & 18.5 & 13.1 \\
% car & 24.2 & 18.0 & 27.9 & 24.4 \\
% tvmonitor & 39.5 & 19.4 & 50.5 & 19.7 \\
% sofa & 25.5 & 18.3 & 31.0 & 24.2 \\
% motorbike & 20.7 & 12.1 & 15.9 & 9.7\\
% diningtable & 11.3 & 11.1 & 40.4 & 8.9 \\
% chair & 11.0 & 6.5 & 11.1 & 7.11\\
% bus & 89.8 & 82.0 & 78.1 & 76.6 \\
% bottle & 21.8 & 10.8 & 5.5 & 3.9 \\
% boat & 24.6 & 21.9 & 30.1 & 27.7\\
% bicycle & 10.1 & 5.3 & 9.8 & 5.8\\
% train & 143.9 & 154.9 & 99.1 & 113.2\\
% \hline
% Mean & 36.6 & 31.1 & 34.8 & 27.9\\
% \hline
%     \end{tabular}
%     \caption{Caption}
%     \label{tab:pascal_per_cat}
% \end{table}

\begin{table}[]
    \centering
    \small
    \begin{tabular}{c|c}
    \hline
        dataset & dictionary sizes \\
        \hline
        CMU-Mocap & 512, 256, 128, 64, 32, 16, 8\\
        UP3D & 512, 256, 128, 64, 32, 16, 8\\
        Pascal3D+ & 256, 128, 64, 32, 16, 8, 4, 2\\
        H3.6M & 125, 115, 104,  94,  83,  73,  62,  52,  41,  31,  20, 10\\
        NRSfM Challenge & 125, 115, 104,  94,  83,  73,  62,  52,  41,  31,  20, 10\\
        Apollo & 128, 100, 64, 50, 32, 16, 8, 4 \\
        \hline
    \end{tabular}
    \caption{Dictionary sizes used in each reported experiment.}
    \label{tab:dict_size}
\end{table}

\begin{table}[h]
    \centering
    \small
    \begin{tabular}{c|cc|cc}
    \hline
    \footnotesize{Method} & \footnotesize{MV/T} & \footnotesize{E3D}  & \scriptsize{MPJPE} & \scriptsize{PA-MPJPE}\\
    \hline
    Martinez \etal~\cite{martinez2017simple} & -& -  & 62.9 & 52.1 \\
    Zhao \etal~\cite{Zhao_2019_CVPR} & - & -  & 57.6 & - \\
    \hline
    3DInterpreter~\cite{w_3dinterp} & & \checkmark  & - & 98.4\\
    AIGN~\cite{w_aign} & & \checkmark  & - & 97.2\\
    Tome \etal ~\cite{Tome_2017_CVPR} & \checkmark & \checkmark & 88.4 & - \\
    RepNet~\cite{repnet} & & \checkmark & 89.9 & 65.1 \\
    Drover \etal~\cite{drover} & & \checkmark & - & 64.6\\
    \hline
    \hline
    Pose-GAN~\cite{posegan} & & & 173.2 & - \\
    C3DPO~\cite{c3dpo} & & & 145.0 & - \\
    Wang \etal ~\cite{Wang_2019_ICCV}  & & & 83.0 & 57.5 \\
    Chen~\etal~\cite{chen_drover} & \checkmark & & - & 68\\
    \hline
    Ours(persp proj) & & & 68.9 & 59.4\\
    + scale corr itr1 & & & 67.3 & 59.2\\
    + scale corr itr2 & & & \textbf{67.0} & \textbf{58.7}\\
    \hline
    \end{tabular}
    \caption{Result on \textbf{H3.6M} dataset with detected 2D kepoint input. In our result, we use detected points from cascaded pyramid network (CPN~\cite{chen2018cascaded}) which is finetuned on H3.6M training set (excluding S9 and S11) by \cite{pavllo20193d}.}
    \label{tab:h36m_cpn}
\end{table}

\begin{table*}[h]
\centering
\small
\begin{tabular}{c|c|c|c|c|c|c}
\hline
    & temporal & Articulated & Balloon & Paper & Stretch & Tearing \\
    \hline
    % Multibody~\cite{kumar2016multi} & 10.15 & 10.64 & 15.78 & 9.96 & 14.17 \\
    % BMM~\cite{dai2014simple} & 24.54 & 12.91 & 22.37 & 18.71 & 18.87\\
    % Kumar~\cite{kumar2020non} & 12.02 & 11.79 & 16.21 & 12.05 & 16.08\\
    % \hline
    % Ours & 45.52 & 3.85 & 13.32 & 4.9 & 9.32 \\
    % \hline
    
    CSF2 & $\checkmark$ & 35.51 & 19.01 & 33.95 & 23.22 & 18.77 \\
    KSTA & $\checkmark$ & 42.11 & 18.45 & 32.18 & 22.88 & 17.59\\
    PTA & $\checkmark$ & 36.71 & 28.88 & 41.72 & 30.45 & 23.14\\
    Bundle & $\checkmark$ & 64.48 & 36.40 & 41.64 & 36.64 &28.73\\
    SoftInext & x & 61.43 &36.75 & 47.41 & 45.56 & 37.87\\
    Compressible & x & 72.77 & 52.53 & 62.44 & 57.45 & 54.71 \\
    MDH & x &  88.66 & 58.27 & 66.98 & 66.27 & 56.67\\
    \hline
    Ours & x & 47.80 & 30.16 & 44.33 & 36.59 & 30.74 \\
    \hline
\end{tabular}
\caption{Comparison on \textbf{NRSfM Challenge Dataset} with \textbf{orthogonal} projection and no missing data.}
\label{tab:chall_orth}
\end{table*}

\begin{table*}[h]
\centering
\small
\begin{tabular}{c|c|c|c|c|c|c|c}
\hline
    &  temporal &Articulated & Balloon & Paper & Stretch & Tearing \\
    \hline
    % Multibody~\cite{kumar2016multi} & 10.15 & 10.64 & 15.78 & 9.96 & 14.17 \\
    % BMM~\cite{dai2014simple} & 24.54 & 12.91 & 22.37 & 18.71 & 18.87\\
    % Kumar~\cite{kumar2020non} & 12.02 & 11.79 & 16.21 & 12.05 & 16.08\\
    % \hline
    % Ours & 45.52 & 3.85 & 13.32 & 4.9 & 9.32 \\
    % \hline
    
    CSF2 & \checkmark & 49.75 & 29.14 & 38.07 & 33.07 & 26.09 \\
    KSTA & \checkmark &44.49 & 27.94 & 36.06 & 29.23 & 23.01\\
    PTA & \checkmark & 58.08 & 37.20 & 47.42 & 41.40 & 30.96\\
    Bundle & \checkmark & 58.95 & 37.07 & 40.88 & 38.16 & 27.40\\
    SoftInext & x &69.11 & 41.76 & 53.91 & 48.98 & 45.93\\
    Compressible & x & 88.00 &  55.21 & 67.44 & 63.11 & 53.93 \\
    MDH & x & 91.55 & 58.00 & 66.54 & 62.49 & 53.93\\
    \hline
    Ours & x & 65.93 & 31.91 & 41.03 & 47.61 & 38.98 \\
    \hline
\end{tabular}
\caption{Comparison on \textbf{NRSfM Challenge Dataset} with \textbf{Perspective} projection and no missing data.}
\label{tab:chall_persp}
\end{table*}